\documentclass[10pt,twocolumn,letterpaper]{article}

\usepackage{3dv}
\usepackage{times}
\usepackage{epsfig}
\usepackage{graphicx}
\usepackage{amsmath}
\usepackage{amssymb}
\usepackage{enumitem}

\usepackage[dvipsnames]{xcolor}
\usepackage{comment}
\usepackage{multirow}
\usepackage{float}

\usepackage{xcolor}
\usepackage{float}
\usepackage{capt-of}
\usepackage{booktabs}
\usepackage{varwidth}

\usepackage{tablefootnote}
\usepackage{multirow}
\usepackage{multicol}
\usepackage{siunitx}
\usepackage{multirow}
\usepackage{tablefootnote}
\usepackage{pdflscape}
\usepackage{pifont}
\usepackage{subcaption} 
\usepackage{booktabs}
\usepackage{makecell}
\usepackage{dsfont}
\usepackage[font=small]{caption}

\newcommand{\cmark}{\ding{51}}%
\newcommand{\xmark}{\ding{55}}%
\threedvfinalcopy 

\def\threedvPaperID{****} 

\ifthreedvfinal\fi
\begin{document}

\title{3D Reconstruction of Novel Object Shapes from Single Images}

\author{Anh Thai\thanks {Equal contribution.}, Stefan Stojanov\footnotemark[1], Vijay Upadhya, James M. Rehg \\
Georgia Institute of Technology\\
\{athai6, sstojanov, vupadhya6, rehg\}@gatech.edu
}

\maketitle

\begin{abstract}
Accurately predicting the 3D shape of any arbitrary object in any pose from a single image is a key goal of computer vision research. This is challenging as it requires a model to learn a representation that can infer both the visible and occluded portions of any object using a limited training set. A training set that covers all possible object shapes is inherently infeasible. Such learning-based approaches are inherently vulnerable to overfitting, and successfully implementing them is a function of both the architecture design and the training approach. We present an extensive investigation of factors specific to architecture design, training, experiment design, and evaluation that influence reconstruction performance and measurement. We show that our proposed SDFNet achieves state-of-the-art performance on seen and unseen shapes relative to existing methods GenRe\cite{genre} and OccNet\cite{mescheder2019occupancy}. We provide the first large-scale evaluation of single image shape reconstruction to unseen objects. The source code, data and trained models can be found on https://github.com/rehg-lab/3DShapeGen.
\vspace{-15pt}
\end{abstract}
\section{Introduction}

\begin{figure}[t!]
\includegraphics[width=\linewidth]{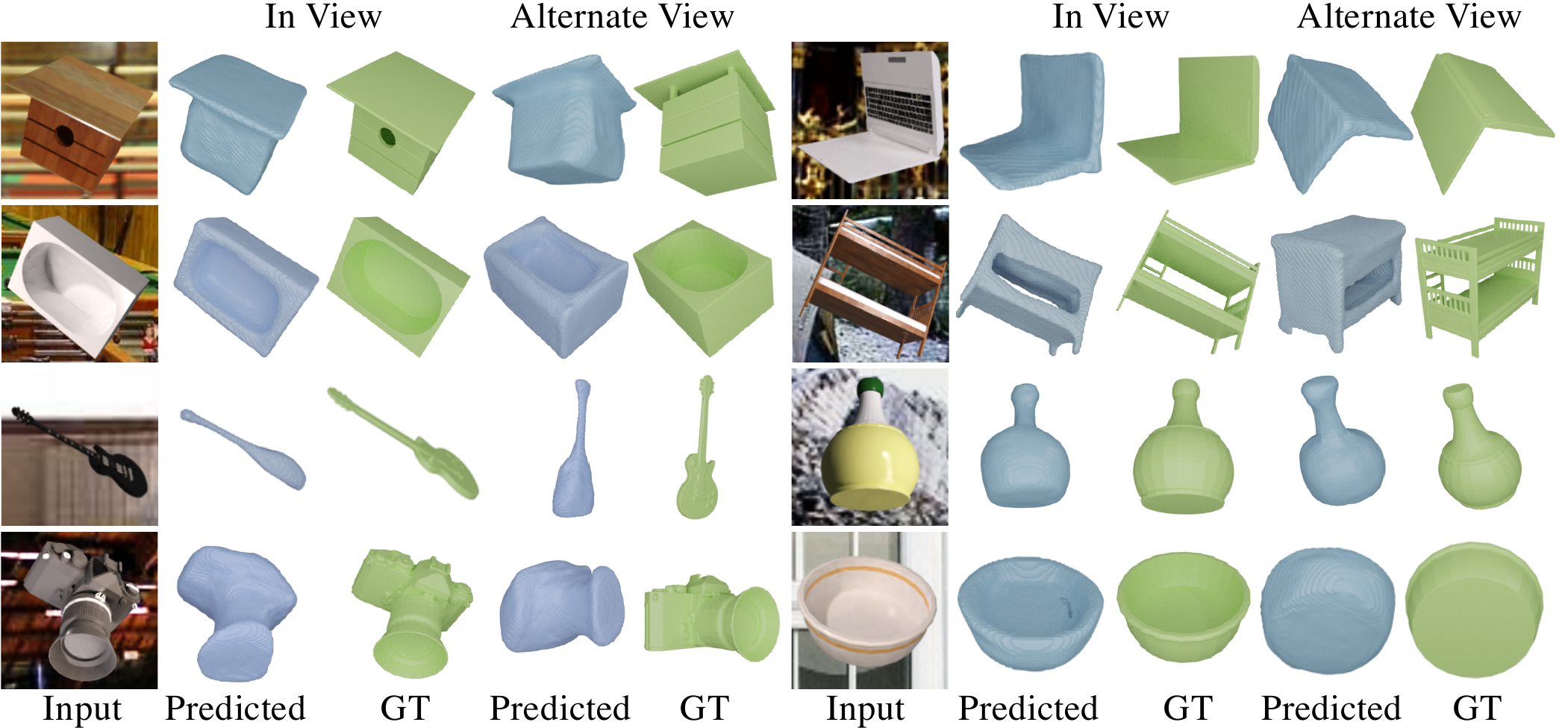}
    \vspace{-15pt}
    \caption{Reconstruction of novel shape classes not seen during training of SDFNet. Reconstructed meshes shown in input view and represenetative alternate view.}
    \label{fig:reconstruction_examples}
\vspace{-10pt}
\end{figure}

The introduction of the large scale 3D model datasets ShapeNet~\cite{chang2015shapenet} and ModelNet~\cite{wu2015modelnet} sparked substantial activity in the development of deep models that can reconstruct the 3D shape of an object from a single RGB image~ \cite{wu20153d,choy20163d,mescheder2019occupancy,xu2019disn,chen2019learning,park2019deepsdf,wang2018pixel2mesh,groueix2018atlasnet,jack2018learning} (see Fig.~\ref{fig:reconstruction_examples} for examples). Surprisingly, in spite of the tremendous progress, basic questions remain unanswered regarding the ability of deep models to \emph{generalize to novel classes of 3D shapes} and the impact of coordinate system and shape representation choices on generalization performance~\cite{tatarchenko2019single,shin2018pixels}. 

The goal of this paper is to provide a comprehensive investigation of the factors that determine generalization performance. We posit that generalization in 3D shape reconstruction requires models to produce accurate 3D shape outputs for unseen \emph{classes} of objects, under \emph{all poses}, and over variations in \emph{scene lighting, surface reflectance, and choice of background}. In the context of classical work on 3D reconstruction in areas such as SFM, these criteria may seem reasonable, but we are the first to apply such strong criteria for generalization to single image 3D shape reconstruction. Our investigation addresses the following four key questions:
\begin{enumerate}[itemsep=-1.2pt]
    \item How does the choice of 3D coordinate system effect generalization?
    \item What is the impact of an intermediate depth and normals representation on reconstruction performance?
    \item How should object models be rendered in order to test performance effectively?
    \item How well do state-of-the-art methods perform in large-scale generalization tests on unseen object categories?
\end{enumerate}

In order to address these questions, it is necessary to choose a representation for 3D shape and a reconstruction architecture. Substantial prior work has identified signed distance fields (SDF) as a superior representation in comparison to surface meshes and discrete and continuous occupancies~\cite{xu2019disn,mescheder2019occupancy, chabra2020deepLS, park2019deepsdf, chen2019learning}.
Continuous SDF representations are efficient\footnote{In comparison to grid representations, SDF require less memory and do not require computationally expensive resampling for geometric  transformations.} and effective in capturing detailed shape features since the SDF encodes the distance to the mesh surface in addition to occupancy. In order to leverage these advantages, we introduce a novel reconstruction architecture, \emph{SDFNet}, which achieves substantial improvements in reconstruction accuracy relative to previous state-of-the-art methods employing both discrete~\cite{genre} and continuous~\cite{mescheder2019occupancy} shape representations. We are the first to incorporate an explicit depth and normal maps (2.5D) estimation module with an implicit continuous shape representation, which makes it feasible to study the above questions more effectively. 
By releasing our SDFNet implementation and evaluation software, we hope to encourage and facilitate continued investigations into large-scale generalization.

The first issue we investigate is the choice of coordinate representation for training. Prior work has shown that using viewer-centered (VC) representations results in improved reconstruction of unseen classes in comparison to an object-centered (OC) approach~\cite{tatarchenko2019single,genre, shin2018pixels,bautista2021generalization}, by discouraging the model from learning to simply retrieve memorized 3D shapes in a canonical pose. We extend this approach by demonstrating that sampling the \emph{camera tilt}, a dimension of variation that was missing in prior works, along with pitch and yaw, has a surprisingly large impact on performance.

The second question addresses the utility of incorporating an explicit 2.5D image representation as an intermediate representation between the RGB image input and the learned latent shape feature vector (used for SDF prediction). An intermediate 2.5D representation forces the model to learn how to encode the visible object shape in the form of depth and normal images. Explicitly encoding the visible object shape is valuable, but is not guaranteed when directly training a CNN-based shape encoding from RGB images. 2.5D representations were explored in the Multi-View~\cite{shin2018pixels} and GenRe~\cite{genre} architectures as a means to facilitate good generalization performance. Our work provides a more extensive assessment of the impact of 2.5D (Sec.~\ref{sec:method_compare}).

The third issue we investigate is the impact of photorealistic object rendering on reconstruction performance. Prior works primarily use Lambertian shading and white backgrounds. We demonstrate, unsurprisingly, that models trained on Lambertian images do not generalize well (Sec.~\ref{sec:rendering}) to images containing complex light sources, surface reflectance, and variations in the background scene (illustrated in Fig~\ref{fig:rendering_effects}). More importantly, we incorporate complex rendering into the training and testing data for all of our experiments, allowing us to validate the effectiveness of SDFNet in generalizing across complex rendering effects. By releasing our rendering framework as part of our project software, we hope to enable the research community to incorporate an additional dimension of generalization.

The fourth issue is the impact of large scale evaluation on unseen categories. Zhang et al.~\cite{genre} proposed testing on \emph{unseen} ShapeNet categories as an effective test of generalization, presenting results using 3 training and 10 testing classes. We extend this approach significantly by presenting the first results to use all of the meshes in ShapeNetCore.v2, training on 13 classes and testing on 42 unseen classes, which contain \emph{two orders of magnitude more meshes} than prior works (Sec.~\ref{sec:dataset}). We also present the first analysis of \emph{cross-dataset generalization} for 3D shape reconstruction, using ShapeNet and ABC~\cite{koch2019abc} (Sec.~\ref{sec:cross_dataset}). This paper makes the following contributions: 
\begin{itemize}[itemsep=-1.2pt]
\item The SDFNet architecture, which combines a 2.5D representation with an SDF shape representation and achieves state-of-the-art performance 
\item Comprehensive evaluation of four key factors affecting reconstruction generalization
\item Introduction of a new large-scale generalization task involving all meshes in ShapeNetCore.v2 and a subset of the ABC~\cite{koch2019abc} dataset.
\end{itemize}

This work is organized as follows: Sec.~\ref{sec:related} presents our work in the context of prior methods for shape reconstruction. Sec.~\ref{sec:method} provides a description of our approach and introduces the SDFNet architecture. It also addresses the consistent evaluation of metrics for surface-based object representations. Sec.~\ref{sec:exp} presents our experimental findings. It includes a discussion of consistent practices for rendering images and a training/testing split for ShapeNetCore.v2 which supports the large-scale evaluation of generalization to unseen object categories.

\begin{table}[t!]
\begin{minipage}[t!]{\linewidth}
    
\begin{center}
\resizebox{\linewidth}{!}{

\begingroup
\setlength{\tabcolsep}{4pt} 
\renewcommand{\arraystretch}{1.2} 

\begin{tabular}{l|c|c|c}
 Method & \begin{tabular}{@{}c@{}}2.5D \vspace{-3pt} \\ Estimator\end{tabular} & \begin{tabular}{@{}c@{}}Continuous \vspace{-3pt} \\ Representation\end{tabular} & \begin{tabular}{@{}c@{}}3-DOF \vspace{-3pt}\\ Viewer-Centered \end{tabular}\\
\hline
GenRe \cite{genre}                     & \cmark & \xmark & \xmark        \\    
Multi-View \cite{shin2018pixels}       & \cmark & \xmark & \xmark        \\
GSIR \cite{wanggsir}                   & \cmark & \xmark & \xmark        \\
\hline
OccNet \cite{mescheder2019occupancy}   & \xmark & \cmark & \xmark           \\
DISN \cite{xu2019disn}                 & \xmark & \cmark & \xmark        \\
3D43D \cite{bautista2021generalization} & \xmark & \cmark & \xmark \\
\hline
SDFNet                              & \cmark & \cmark & \cmark          \\
\hline
\end{tabular}


\endgroup
}

\end{center}

    \vspace{-15pt}
    \caption{Our proposed SDFNet architecture has a unique set of characteristics compared to prior single-view 3D reconstruction methods.}
    \label{table:method_characteristics}
\end{minipage}%
\vspace{-15pt}
\end{table}

\section{Related Work}
\label{sec:related}

There are two sets of related works on single image 3D reconstruction which are the closest to our approach. The first set of methods were the first to employ continuous implicit surface representations~\cite{mescheder2019occupancy,xu2019disn}. Mescheder et al.~\cite{mescheder2019occupancy} utilized continuous occupancy functions, while Xu et al.~\cite{xu2019disn} shares our use of signed distance functions (SDFs). We differ from these works in our use of depth and normal (2.5D sketch) intermediate representations, along with other differences discussed below. The second set of related works pioneered the use of unseen shape categories as a means to investigate the generalization properties of shape reconstruction~\cite{genre,shin2018pixels,wanggsir, bautista2021generalization}. In contrast to SDFNet, these methods train and test on a small subset of ShapeNetCore.v2 classes, and other than~\cite{bautista2021generalization} they utilize discrete object shape representations. We share with~\cite{genre,shin2018pixels,wanggsir} the use of depth and normal intermediate representations. Bautista et al.~\cite{bautista2021generalization} evaluate generalization to unseen categories and use continuous occupancies and emphasize reconstruction from multiple views. Additionally, we differ from all prior works in our choice of object coordinate representation (VC with 3-DOF, Sec.~\ref{sec:method}), in the scale of our experimental evaluation (using all 55 classes of ShapeNetCore.v2 to test generalization on seen and unseen classes, Sec.~\ref{sec:dataset}), and our investigation into the effects of rendering in Sec.~\ref{sec:rendering}. We provide direct experimental comparison to GenRe~\cite{genre} and OccNet~\cite{mescheder2019occupancy} in Sec.~\ref{sec:method_compare}, and demonstrate the improved performance of SDFNet. GenRe~\cite{genre}, GSIR\cite{wanggsir}, 3D43D~\cite{bautista2021generalization} and DISN~\cite{xu2019disn} require known intrinsic camera parameters in order to perform projection. In contrast, SDFNet does not require projection, and only regresses camera translation for estimating absolute depth maps. See Tbl.~\ref{table:method_characteristics} for a summary of the relationship between these prior works and SDFNet. 

Other works also perform 3D shape reconstruction \cite{wu20153d,choy20163d,wang2018pixel2mesh, groueix2018atlasnet, jack2018learning,smith19a,henderson2019learning,Liu_2019_ICCV,liu2019learning} but differ from this work since they perform evaluation on categories seen during training, on small scale datasets, and use different shape representations such as voxels \cite{wu20153d, choy20163d}, meshes \cite{wang2018pixel2mesh,groueix2018atlasnet,jack2018learning,Liu_2019_ICCV}, neural radiance fields~\cite{mildenhall2020nerf} and continuous implicit representations with \emph{2D} supervision \cite{liu2019learning}. Prior works on seen categories used the 13 largest ShapeNet categories, while \cite{tatarchenko2019single} uses all of them. We use all 55 ShapeNet categories, but train on 13 and test on 42.

Partly relevant are works on object viewpoint estimation~\cite{su2015multi}, which advocates for synthetic data rendering with high variability for that task, and recent 6DOF pose estimation works~\cite{su2015multi, banani2020novel} which perform evaluation on unseen shape classes. In comparison, our work focuses on 3D shape reconstruction.


\begin{figure*}[ht!]
\centering
\includegraphics[width=0.85\linewidth]{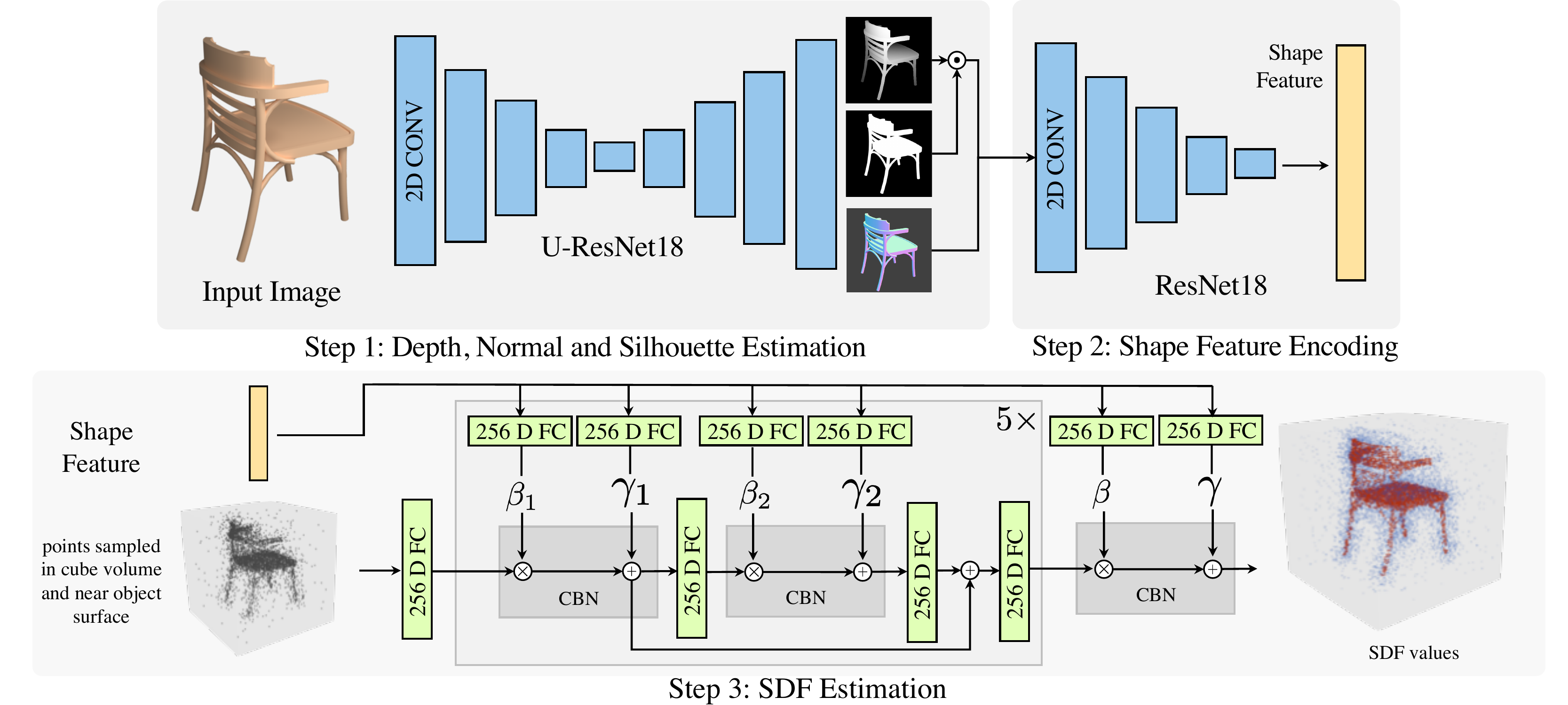}
\vspace{-10pt}
\caption{Two-stage SDFNet architecture: 2.5D estimation and 3D shape completion. The 2.5D sketch estimator is a U-ResNet18 \cite{ronneberger2015unet} architecture as in \cite{genre}. Given the depth and surface normal output, the 3D completion module produces a feature encoding used to produce conditional batch norm (CBN \cite{de2017modulating}) values for an MLP as in OccNet\cite{mescheder2019occupancy} and assigns SDF values to a sampled set of points in 3D.}
\label{fig:arch}
\vspace{-17pt}
\end{figure*}


\emph{2.5D Sketch Estimation}. We use the 2.5D sketch estimator module from MarrNet~\cite{wu2017marrnet} to generate depth and surface normal maps as intermediate representations for 3D shape prediction.  We note that there is a substantial body of work on recovering intrinsic images~\cite{barrow1978recovering,ikeuchi1981numerical,facil2019cam,chen2016single,kuznietsov2017semi,Li_2018_CVPR, tappen2003recovering} which infers the properties of the visible surfaces of objects, whileour focus is additionally on hallucinating the self-occluded surfaces and recovering shape for unseen categories. More recent works use deep learning in estimating depth \cite{facil2019cam,chen2016single,kuznietsov2017semi,Li_2018_CVPR}, surface normals \cite{huang2019framenet,Zeng2019DeepSN}, joint estimation \cite{Qi_2018_CVPR,Qiu_2019_CVPR}, and other intrinsic images~\cite{li2018learning,Li2018CGIntrinsicsBI,baslamisli2018cnn}.

\emph{Generative Shape Modeling}. This class of works \cite{park2019deepsdf,chen2019learning, kleineberg2020adversarial} is similar to SDFNet in the choice of shape representation but primarily focuses on learning latent spaces that enable shape generation. IM-NET~\cite{chen2019learning} contains limited experiments for single view reconstruction, done for one object category at a time, whereas DeepSDF~\cite{park2019deepsdf} and DeepLS~\cite{chabra2020deepLS} also investigate shape completion from partial point clouds.

\emph{3D Shape Completion}. This class of works \cite{Song_2017_CVPR,Firman_2016_CVPR,Rock_2015_CVPR,Yang_2017_ICCV,stutz2018learning, Giancola2019LeveragingSC, shin2018pixels} is not directly related to our primary task because we focus on 3D shape reconstruction from single images. Note that in Secs.~\ref{sec:viewer_centered} and~\ref{sec:cross_dataset} we utilize ground truth single-view 2.5D images as inputs, similar to these prior works. 

\section{Method}\label{sec:method}

In this section we introduce our~\emph{SDFNet} architecture, illustrated in Fig.~\ref{fig:arch}, for single-view 3D object reconstruction based on signed distance fields (SDF). We describe our architectural choices along with the design of a 3-DOF approach to viewer-centered object representation learning that improves generalization performance. We are the first to train models to learn a 3-DOF representation, using a novel architecture for 3D shape reconstruction (see Tbl.~\ref{table:method_characteristics}). Our approach achieves SOTA performance in single-view 3D shape reconstruction from novel categories.

\subsection{SDFNet Architecture} Our deep learning architecture, illustrated in Fig.~\ref{fig:arch}, incorporates three main elements: 1) A 2.5D estimation module that produces depth and normal map estimates from a single RGB input image, followed by 2) a CNN-based object shape feature encoder that produces a shape feature vector from the predicted 2.5D, and 3) a continuous shape estimator that learns to regress an implicit 3D surface representation based on signed distance fields (SDF) from the learned feature encoding of the 2.5D sketch. 

The use of depth and normals as intermediate representations is motivated by the observation that such intermediate intrinsic images explicitly capture surface orientation and displacement, key object shape information. As a result, the shape reconstruction module is not required to jointly learn both low-level surface properties and the global attributes of 3D shape within a single architecture. Prior works that study generalization and domain adaptation in 3D reconstruction~\cite{genre,wu2017marrnet} have also adopted these intermediate representations and have demonstrated their utility in performing depth estimation for novel classes. Note that, following the approach in \cite{genre}, we use ground truth silhouettes when converting from a normalized depth map to an absolute depth map. Our findings are that models trained with an intermediate 2.5D representation have a slight reconstruction performance improvement in comparison to models trained directly from images, but are more robust to unseen variations in lighting, object surface reflection an backgrounds (for further discussion see Secs.~\ref{sec:method_compare} and~\ref{sec:rendering}).

As in~\cite{mescheder2019occupancy} we adopt an implicit shape regression module to predict 3D object shape. Unlike~\cite{mescheder2019occupancy}, our approach regresses SDF values from the object surface and is learned using a different procedure from~\cite{mescheder2019occupancy}. Our viewer-centered approach is designed to allow for learning general implicit shape functions (detailed in the following section Sec.~\ref{subsec:learning-general-implicit-3D}).

\subsection{Learning General Implicit 3D Object Shape Representations}
\label{subsec:learning-general-implicit-3D}

A basic question underlies all 3D shape reconstruction approaches: How many rotational degrees of freedom of should the learned shape representation encode? A commonly used approach~\cite{choy20163d, tatarchenko2016multi, groueix2018atlasnet} is to train the model to encode no such degrees of freedom. Such methods, known as learning \emph{object centered} (OC) representations, are trained to always predict object shape in a pre-defined, \textit{canonical pose} (generally consistent across an object category), despite variations in the object pose in the input images. Recent works~\cite{genre,shin2018pixels,tatarchenko2019single} have argued that training shape prediction models in a way that allows for predicting object shape with 2 rotational degrees of freedom, azimuth and elevation, prevents them from performing reconstruction in a recognition regime (i.e. memorizing training shapes in canonical pose), resulting in more effective generalization. These models learn representations known as 2-DOF viewer-centered (2-DOF VC) representations.

2-DOF VC representations have a significant disadvantage: the set of object views remains biased to the canonical pose for each category, hindering the generalization ability of the learned representation to images of novel object categories and objects presented in arbitrary poses. This is a result of the vertical axis of object models in the same category being aligned with the same gravity reference. Our proposed solution is to add camera tilt along with rotation and elevation to the training data, significantly increasing the variability in the training and testing data. We refer to this as 3-DOF VC (see Fig.~\ref{fig:rendering_effects}). Our work is also the first to investigate learning viewer centered representations using implicit shape functions. Implicit shape methods are particularly well suited for viewer centered training, since generating rotated ground truth points can be done with a single matrix multiplication, rather than the computationally inefficient rotating and resampling of dense voxel grids. We demonstrate in Sec.~\ref{sec:viewer_centered} that the 3-DOF VC approach results in better generalization performance in comparison to both OC and 2-DOF VC object representations.

\begin{figure*}[t!]
\centering
\includegraphics[width=0.75\linewidth]{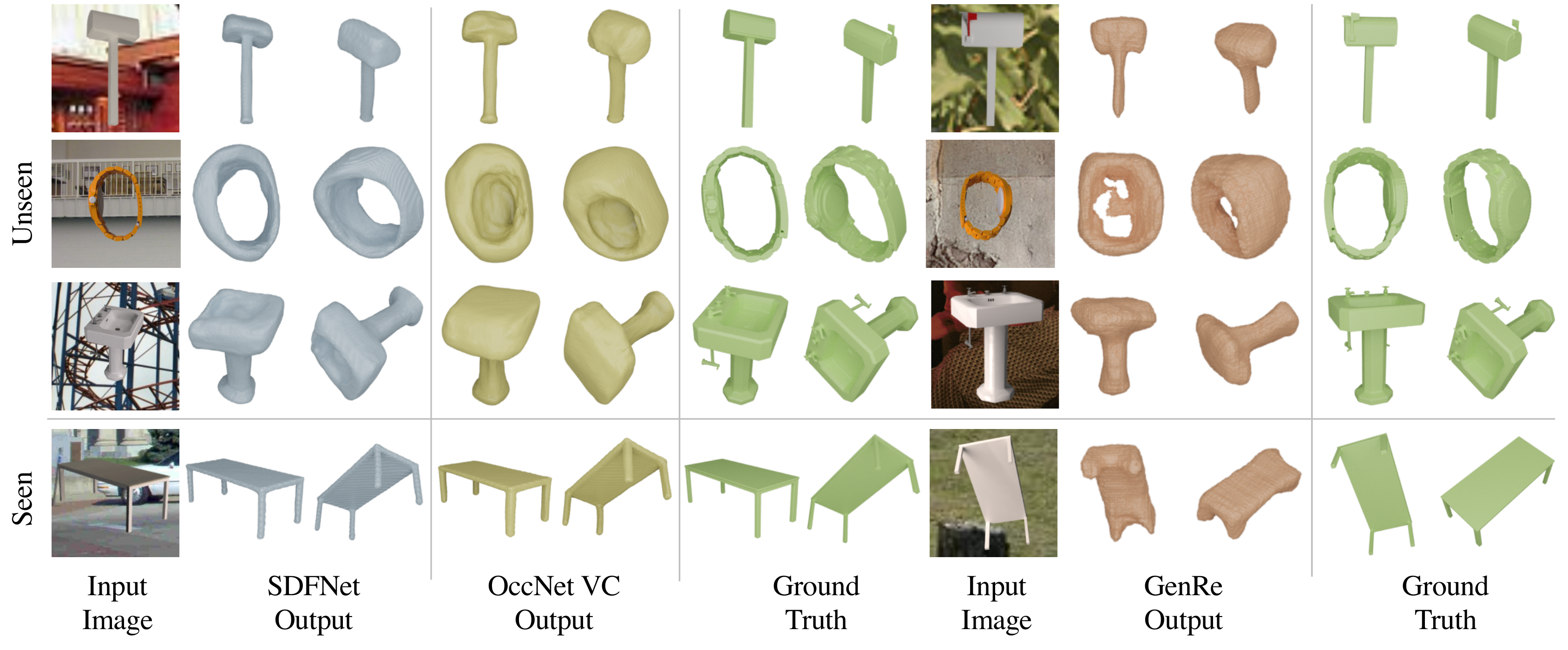}
\vspace{-10pt}
\caption{Qualitative results of SDFNet, OccNet VC and GenRe on seen and unseen classes of 2-DOF viewpoint ShapeNetCore.v2 testing data with LRBg renderings. Each column shows the results from a method in two different views: input view (left), other view (right). Note that we show GenRe's performance on different renders of the same set of objects with the same rendering settings. See Supplement for further discussion.}
\label{fig:genre_occnet_sdfnet}
\vspace{-17pt}
\end{figure*}

\textbf{SDF Points Sampling Strategy} We generate ground truth SDF values for points densely sampled close to the mesh surface. For details refer to the Supplement. 

\textbf{Optimization Procedure} We first train the 2.5D estimator using MSE loss between the predicted and ground truth 2.5D representations. In the second stage, we train the CNN feature extractor and SDF estimator by optimizing the $L_1$ loss between the ground truth SDF and the predicted values of the input 3D points. Training is done using the Adam optimizer~\cite{kingma2014adam} with $lr=1e^{-4}$.

\section{Experiments}
\label{sec:exp}

This section presents the results of our large-scale experimental evaluation of SDFNet and related methods. It is organized as follows: Our choice of datasets and train/test splits is outlined in Sec.~\ref{sec:dataset}, followed by a discussion of metrics for evaluation in Sec.~\ref{sec:metric}. Sec.~\ref{sec:method_compare} reports on the generalization performance of SDFNet, GenRe~\cite{genre}, and OccNet~\cite{mescheder2019occupancy}. In Sec.~\ref{sec:2.5D_eval} we investigate the utility using 2.5D as an intermediate representation. In Sec.~\ref{sec:viewer_centered}, we investigate the impact of the choice of object coordinate representation on generalization ability, while Sec.~\ref{sec:rendering} discusses the impact of the image rendering process. Last, in Sec.~\ref{sec:cross_dataset} we analyze cross-dataset generalization for 3D shape reconstruction. Unless otherwise specified, the images used were rendered with light and reflectance variability, superimposed on random backgrounds from SUN Scenes~\cite{xiao2010sun}, as described in Sec.~\ref{sec:rendering} (the LRBg condition).

For all of the experiments, we train on one random view per object for each epoch. Testing is done on one random view per object in the test set. We find that the standard deviation for all metrics is approximately $10^{-4}$ based on three evaluation runs.

\subsection{Datasets, Evaluation Split, and Large Scale Generalization}\label{sec:dataset}

\textbf{Datasets} Our experiments use all 55 categories of ShapeNetCore.v2~\cite{chang2015shapenet}. Additionally, in Sec.~\ref{sec:cross_dataset} we use a subset of 30,000 meshes from ABC~\cite{koch2019abc}. To generate images, depth maps, and surface normals, we implemented a custom data generation pipeline in Blender \cite{blender} using the Cycles ray tracing engine. Our pipeline is GPU-based, supports light variability---point and area sources of varying temperature and intensity, and includes functionality that allows for specular object shading of datasets such as ShapeNet, which are by default diffuse.

\textbf{Data Generation} In our experiments, we use 25 images per object in various settings as illustrated in Fig.~\ref{fig:rendering_effects} resulting in over 1.3M images in total. In contrast with prior approaches to rendering 3D meshes for learning object shape \cite{tatarchenko2019single, shin2018pixels, choy20163d}, we use a ray-tracing renderer and study the impact of lighting and specular effects on performance. The ground truth SDF generating procedure is described in Sec.~\ref{sec:method}. We convert SDF values to mesh occupancy values by masking: $\mathds{1}\{\text{sdf}\leq i\}$ where $i$ is the isosurface value.

\textbf{Data Split}  We use the 13 largest object categories of ShapeNetCore.v2 for training as \emph{seen} categories, and the remaining 42 categories as \emph{unseen}, testing data (not used at all during training and validation).
For ABC, we use 20K for training, 7.5K for testing, and 2.5K for validation.

\textbf{Scaling up Generalization} Our generalization experiments are the largest scale to date, using all of the available objects from ShapeNetCore.v2. Our testing set of unseen classes consists of 12K meshes from 42 classes, in comparison to the 330 meshes from 10 classes used in~\cite{genre}.

\subsection{Metrics}
\label{sec:metric}

Following~\cite{tatarchenko2019single}, we use F-Score with percentage distance threshold $d$ (FS@$d$) as the primary shape metric in our experiments, due to its superior sensitivity. We also report the standard metrics IoU, NC (normal consistency) and CD (chamfer distance), which are broadly-utilized despite their known weaknesses (see Fig.~\ref{fig:metric_sequence}.)

A significant practical issue with metric evaluation which has not been discussed in prior work is what we refer to as the \emph{sampling floor} issue. It arises with metrics, such as CD, NC and FS, that require surface correspondences between ground truth and predicted meshes. Correspondences are established by uniformly sampling points from both mesh surfaces and applying    nearest neighbor (NN) point matching. Note that the accuracy of NN matching is a function of the number of sampled points. For a given dataset and fixed number of samples, there is a corresponding sampling floor, which is a bound on the possible error when comparing shapes. We note that sampling 10K points (as suggested in \cite{tatarchenko2019single}) and comparing identical shapes using FS@1 in ShapeNetCore.v2 results in an average sampling floor of 0.8, which admits the possibility of significant error in shape comparisons (since comparing two identical meshes should result in an FS of 1). In our experiments, we use FS@1 and sample 100K points, and have verified that the sampling floor is insignificant in this case. See the Supplement for additional analysis and discussion.

\subsection{SDFNet Outperforms Prior Work at Novel Shape Reconstruction }\label{sec:method_compare}

\begin{table}
\begin{minipage}[t]{\linewidth}
    \begin{center}

\resizebox{\linewidth}{!}{

\begingroup
\setlength{\tabcolsep}{1.85pt} 
\renewcommand{\arraystretch}{1.4} 

\begin{tabular}{l|c|c|c|c|c|c|c|c|c|c}
  & Bch & Vsl & Rfl & Sfa & Tbl & Phn & Spk & Lmp & Dsp & Avg\\
\hline
OccNet & 0.22 & 0.13 & \textbf{0.11} & 0.15 & 0.27 & 0.27 & 0.20 & 0.34 & 0.33 & 0.22\\
GenRe & 0.13 & 0.18 & 0.15 & 0.12 & 0.16 & \textbf{0.14} & 0.18 & 0.19 & 0.21 & 0.16\\
SDFNet & \textbf{0.09} & \textbf{0.12} & 0.15 & \textbf{0.08} & \textbf{0.09} & \textbf{0.14} & \textbf{0.10} & \textbf{0.17} & \textbf{0.17} & \textbf{0.12} \\
\hline
\end{tabular}

\endgroup}
\end{center}

    \vspace{-10pt}
    \caption{SDFNet has the best performance overall on 9 unseen classes of the data split released by GenRe. We are reporting CD since this is the sole evaluation metric utilized by GenRe.}
    \label{table:method_comparison_genre_data}
\end{minipage}%
\vspace{-10pt}

\end{table}


We evaluate the generalization performance of SDFNet relative to two prior works: GenRe~\cite{genre} and OccNet~\cite{mescheder2019occupancy}. GenRe defines the state-of-the-art in single-image object reconstruction for unseen object categories, while OccNet is representative of recent works that use continuous implicit shape representations for shape reconstruction. We cannot reproduce the results of \cite{wanggsir} since they have not released the code and \cite{xu2019disn} does not converge when training on a different data split. We first perform a comparison on the data released by GenRe (3 training categories and 10 unseen testing categories). The result is shown in Tbl.~\ref{table:method_comparison_genre_data}. Our findings indicate that SDFNet outperforms GenRe and OccNet on most of the unseen classes. We then evaluate on a larger scale as described in Sec.~\ref{sec:method}. For this comparison, we use 2-DOF VC data (see Sec.~\ref{sec:method}) to compare with these baseline methods. While GenRe was designed for 2-DOF VC data, OccNet was designed for OC data and was adapted to facilitate a direct comparison. We refer to the adapted model as OccNet VC. 
We note that performing these experiments with GenRe required significant reimplementation effort in order to generate ground truth spherical maps and voxel grids for a large number of additional ShapeNetCore.v2 models (see Supplement for the details). All three parts of GenRe: depth estimation, spherical inpainting, and voxel refinement, are trained using the code base provided by the authors until the loss converges (i.e. 100 epochs without improvement in validation loss.)

Our findings in Tbl.~\ref{table:method_comparison} demonstrate the superior generalization performance of SDFNet relative to OccNet VC and GenRe. Compared to OccNet VC, SDFNet performs better for CD and FS@1, which can be interpreted as an improved ability to capture the thin details of shapes as a result of the better-defined isosurface of the SDF representation. The performance difference relative to GenRe shows the advantage of using a continuous implicit representation. These findings further suggest that good generalization performance can be achieved without explicit data imputation procedures, such as the spherical inpainting used in GenRe. Note that the performance gaps between SDFNet and the baselines are significantly more noticeable in Tbl.~\ref{table:method_comparison} than in Tbl.~\ref{table:method_comparison_genre_data}, indicating the importance of the large scale evaluation. Qualitative results are shown in Fig.~\ref{fig:genre_occnet_sdfnet}. SDFNet demonstrates the ability to capture concavity better than OccNet VC and GenRe in the sink (third row) and the hole of the watch (second row). Due to the strong performance of SDFNet on generalization over baselines, in the following sections we only conduct experiments on SDFNet.


\begin{table}
\begin{minipage}[t!]{\linewidth}
    \begin{center}

\resizebox{\linewidth}{!}{

\begingroup
\setlength{\tabcolsep}{1.85pt} 
\renewcommand{\arraystretch}{1.4} 

\begin{tabular}{l|ccc|c|ccc|c}
&\multicolumn{4}{c|}{Seen Classes}& \multicolumn{4}{c}{Unseen Classes} \\
\hline
 Method & CD & IoU & NC & FS@1 & CD & IoU & NC & FS@1\\
\hline
GenRe               & 0.153 & N/A & 0.60 & 0.12 & 0.172 & N/A & 0.61 & 0.11 \\
OccNet VC            & 0.078 & \textbf{0.72} & 0.78 & 0.27 & 0.11 & \textbf{0.67} & \textbf{0.76} & 0.22\\
SDFNet Img          & 0.058 & 0.70 & 0.78 & 0.36 & \textbf{0.08} & 0.64 & \textbf{0.76} & 0.28\\
SDFNet               & \textbf{0.05} & \textbf{0.72} & \textbf{0.79} & \textbf{0.42} & \textbf{0.08} & 0.66 & \textbf{0.76} & \textbf{0.31}\\ 

\hline

\end{tabular}

\endgroup}
\end{center}
    \vspace{-10pt}
    \caption{SDFNet Img (input to the shape encoder is RGB image) outperforms OccNet while SDFNet with intermediate 2.5D representation has the best performance overall. IoU is omitted for GenRe as per the authors' recommendation since the meshes are not guaranteed to be watertight.}
    \label{table:method_comparison}
\end{minipage}%
\vspace{-10pt}
\end{table}


\begin{table}
\begin{minipage}[t]{\linewidth}
    \begin{center}

\resizebox{\linewidth}{!}{
\begingroup
\setlength{\tabcolsep}{2pt} 
\renewcommand{\arraystretch}{1.4} 

\begin{tabular}{l|ccc|c|ccc|c}
&\multicolumn{4}{c|}{Seen Classes}& \multicolumn{4}{c}{Unseen Classes} \\
\hline
 Method & CD & IoU & NC & FS@1 & CD & IoU & NC & FS@1 \\
\hline

SDFNet Img         & 0.088 & 0.65 & 0.73 & 0.25 & 0.10 & 0.61 & \textbf{0.74} & 0.23\\
SDFNet Est        & \textbf{0.082} & \textbf{0.68} & \textbf{0.75} & \textbf{0.29} & \textbf{0.099} & \textbf{0.64} & \textbf{0.74} & \textbf{0.26}\\  
\hline

SDFNet Orcl        & 0.041 & 0.77 & 0.81 & 0.51 & 0.044 & 0.75 & 0.83 & 0.51 \\


\hline
\end{tabular}
\endgroup
}

\end{center}
    \vspace{-15pt}
    \caption{3 versions of SDFNet that differ only in the shape encoder input: Image only (SDFNet Img), Estimated 2.5D (SDFNet Est), and ground truth 2.5D input (SDFNet Orcl). 3-DOF VC data is used. Incorporating an explicit 2.5D representation (SDFNet Est) improves performance.}
    \label{table:25D_experiment}
\end{minipage}
\vspace{-18pt}
\end{table}

\subsection{Intermediate 2.5D Representations Improve Reconstruction Performance}\label{sec:2.5D_eval}

We perform an experiment to evaluate the effect of SDFNet's intermediate representation (estimated surface depth and normals) on reconstruction performance. The results are reported in Tbl.~\ref{table:25D_experiment}. Three SDFNet models were trained: with only image input to the CNN feature extractor (no depth and normals, the SDFNet Img case), with estimated depth and normals as (the standard case, SDFNet Est), and with ground truth depth and normals (the oracle case, SDFNet Orcl). Our findings show an improvement when using an intermediate representation consisting of surface depth and normals rather than regressing SDF from images directly. The result for SDFNet Oracle demonstrates that there is room for significant gains in performance by improving the accuracy of the depth and normal estimator. Note that all remaining subsections (below) are focused on additional evaluations of SDFNet.

\begin{figure*}

\begin{minipage}{.38\linewidth}
\centering

\begin{minipage}{0.9\linewidth}
\centering
\includegraphics[width=0.85\linewidth]{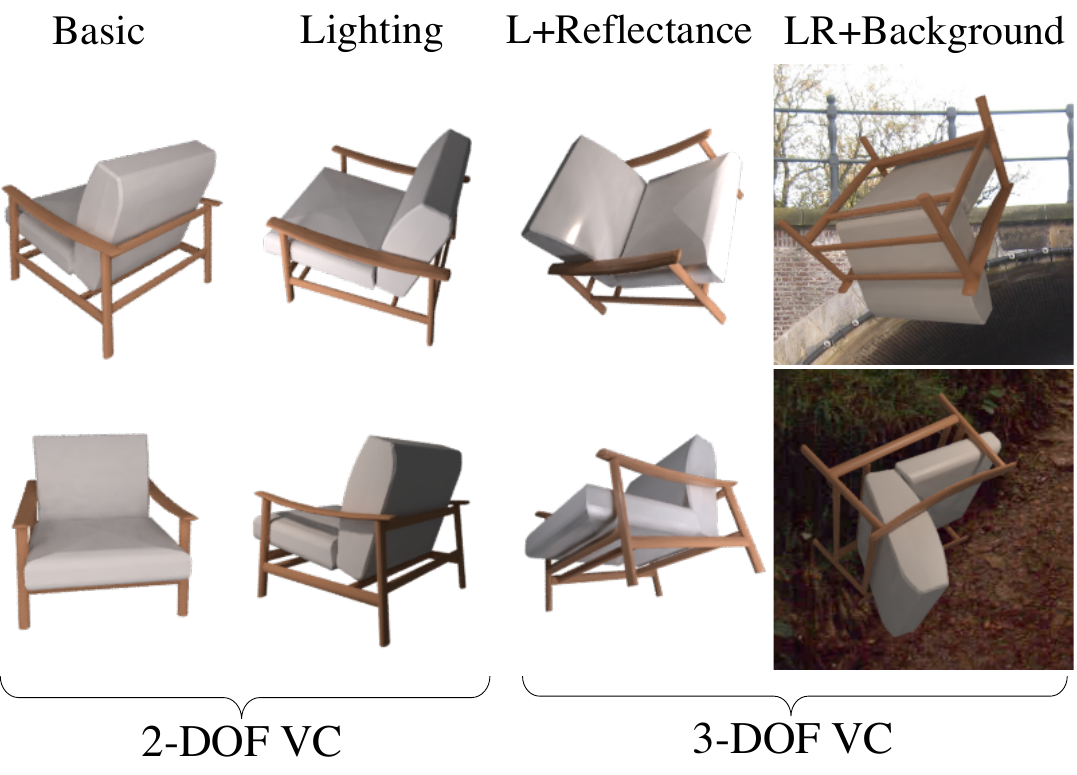}
\vspace{-10pt}
\caption{Data variability in terms of pose, lighting and background.}
\label{fig:rendering_effects}
\end{minipage}

\begin{minipage}{0.9\linewidth}
\centering
\includegraphics[ width=0.9\linewidth]{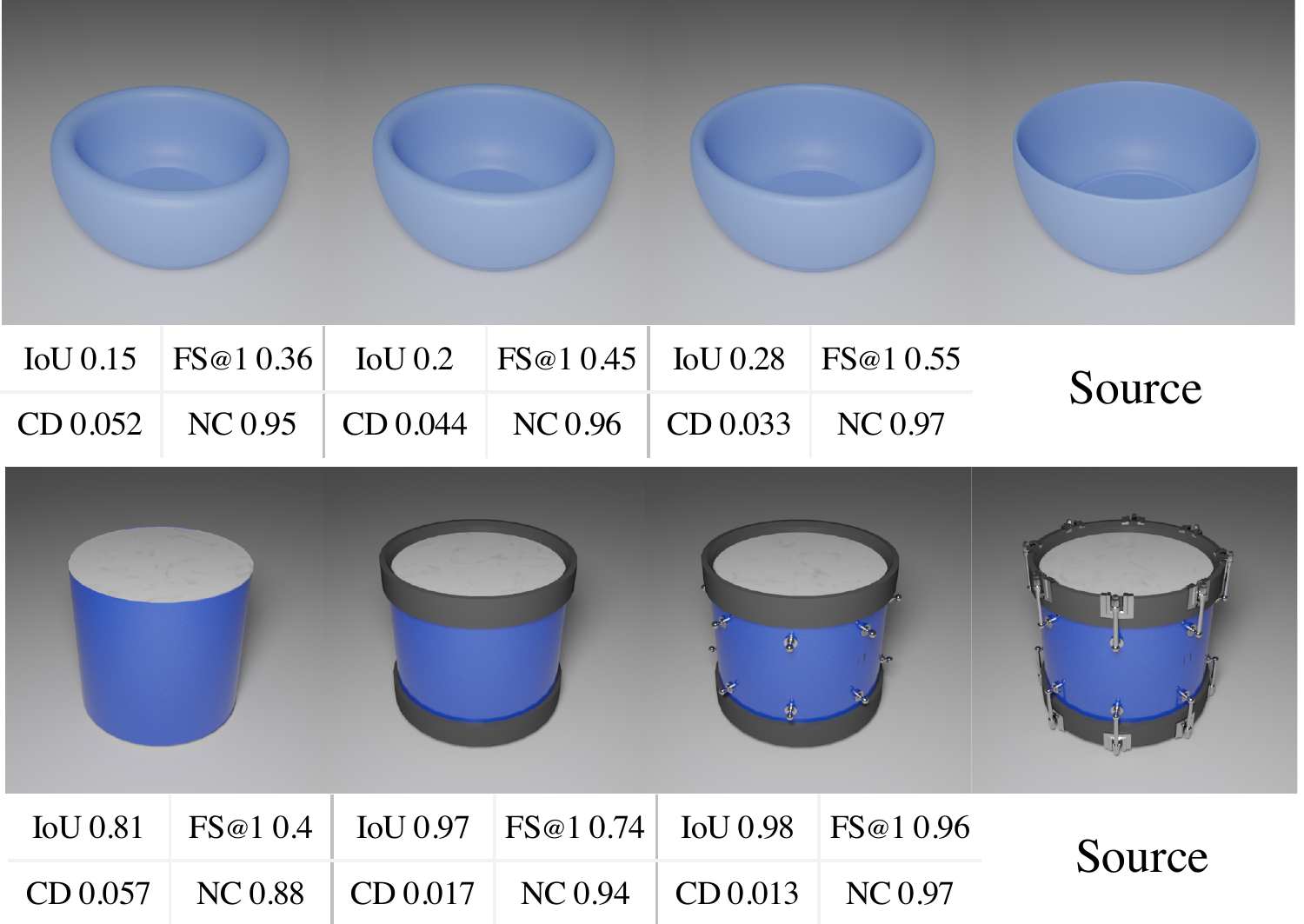}
\vspace{-10pt}
\caption{Significant changes in metrics despite small shape changes.}
\label{fig:metric_sequence}
\vspace{-16pt}
\end{minipage}
\end{minipage}
\hspace{5pt}
\scalebox{1.05}{
\centering
\begin{minipage}{0.55\linewidth}
\centering
\begin{table}[H]
    \begin{center}

\resizebox{1.025\linewidth}{!}{

\begingroup
\setlength{\tabcolsep}{4pt} 
\renewcommand{\arraystretch}{1.2} 

\begin{tabular}{cc|ccc|c|ccc|c}
\multicolumn{2}{c}{ }&\multicolumn{4}{c}{Seen Classes}& \multicolumn{4}{c}{Unseen Classes}\\
\hline
Shape Rep. & Test Data & CD $\downarrow$ & IoU $\uparrow$ & NC $\uparrow$ & FS@1 $\uparrow$  & CD $\downarrow$ & IoU $\uparrow$ & NC $\uparrow$ & FS@1 $\uparrow$\\
\hline
\multirow{2}{*}{OC}         & 2-DOF & 0.040 & 0.71 & 0.82 & 0.51 & 0.093 & 0.58 & 0.74 & 0.33 \\
                            & 3-DOF & 0.134 & 0.51 & 0.66 & 0.21 & 0.189 & 0.40 & 0.60 & 0.14\\
\hline
\multicolumn{2}{c|}{Average}       & 0.087 & 0.61 & 0.74 & 0.36 & 0.141 & 0.49 & 0.67 & 0.23 \\
\hline
\hline
\multirow{2}{*}{2-DOF VC} & 2-DOF & 0.033 & 0.77 & 0.82 & 0.57 & 0.040 & 0.74 & 0.83 & 0.51 \\
                          & 3-DOF & 0.064 & 0.70 & 0.76 & 0.33 & 0.054 & 0.70 & 0.81 & 0.38\\
\hline
\multicolumn{2}{c|}{Average}       & 0.0485 & 0.73 & 0.79 & 0.45 & 0.047 & 0.72  & 0.82 & 0.445 \\
\hline
\hline
\multirow{2}{*}{3-DOF VC} & 2-DOF & 0.038 & 0.77 & 0.82 & 0.53 & 0.041 & 0.75 & 0.84 & 0.52 \\
                          & 3-DOF & 0.041 & 0.77 & 0.81 & 0.51 & 0.044 & 0.75 & 0.83 & 0.51 \\
\hline
\multicolumn{2}{c|}{Average}       & \textbf{0.04} & \textbf{0.77} & \textbf{0.815} & \textbf{0.52} & \textbf{0.042} & \textbf{0.75} & \textbf{0.835} & \textbf{0.515}\\
\hline
\hline 
3-DOF VC & \multirow{2}{*}{3-DOF} & \multirow{2}{*}{0.036} & \multirow{2}{*}{0.81} & \multirow{2}{*}{0.82} & \multirow{2}{*}{0.53} & \multirow{2}{*}{0.038} & \multirow{2}{*}{0.80} &   \multirow{2}{*}{0.84} & \multirow{2}{*}{0.52} \\
55 classes & & & & & & & &\\
\hline
\end{tabular}
\endgroup
}
\end{center}

    \caption{Different coordinate system representations of SDFNet on seen/unseen categories with 2-DOF and 3-DOF viewpoint testing data of ShapeNet. Bold indicates best average performance. Ground truth 2.5D is used, where OC and 2-DOF VC are trained on 2-DOF viewpoint data and 3-DOF VC is trained on 3-DOF viewpoint data. 3-DOF VC has the best reconstruction performance.}
    \label{table:representation_choice}
\end{table}
\end{minipage}
}
\end{figure*}

\subsection{Viewer-Centered Representations Improve Generalization}\label{sec:viewer_centered}

\begin{figure}
\vspace{-5pt}
\begin{minipage}[t!]{\linewidth}
    \includegraphics[width=\linewidth]{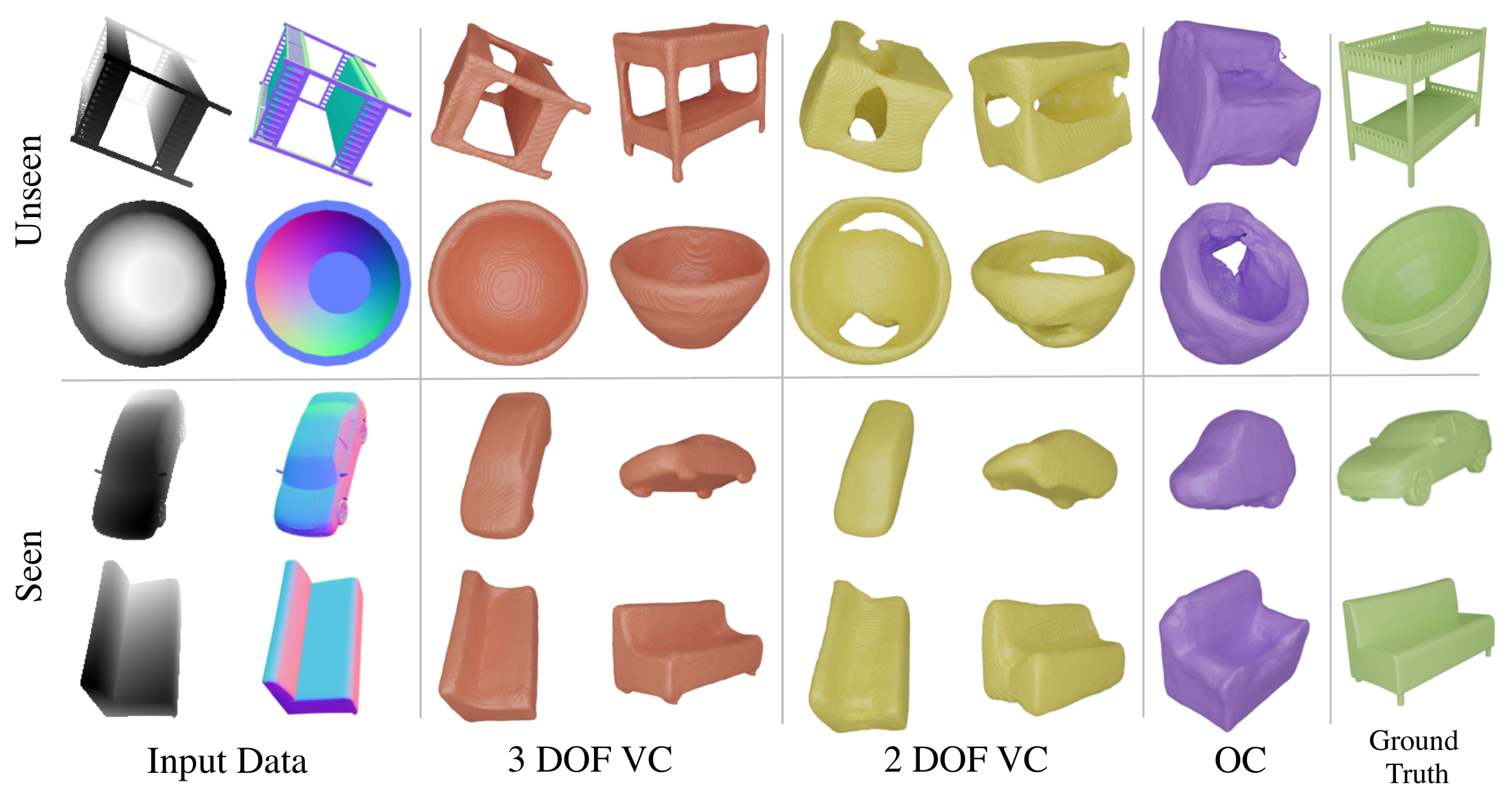}
    \caption{Output meshes from seen and unseen categories with 3-DOF viewpoint testing data of ShapeNetCore.v2.}
    \label{fig:representation_choice}
\end{minipage} 
\vspace{-20pt}
\end{figure}

In this section, we study the effect of the object coordinate representation on reconstruction performance, using SDFNet with ground truth depth and surface normal images as inputs. Three different SDFNet models are trained using object centered (OC), 2-DOF viewer centered (2-DOF VC) and 3-DOF viewer centered (3-DOF VC) representations, as described in Sec.~\ref{sec:method}. OC and 2-DOF VC are trained on 2-DOF viewpoint data and 3-DOF VC is trained on 3-DOF viewpoint data. For successful generalization, it is essential that the model can correctly predict seen and novel object shapes in any arbitrary pose (3-DOF) in addition to poses so that the vertical axis is aligned with gravity (2-DOF). Therefore we evaluate the reconstruction performance on both 2-DOF VC and 3-DOF VC testing data and take the average and present results in Tbl.~\ref{table:representation_choice} and Fig.~\ref{fig:representation_choice}. The 3-DOF VC model performs the best overall on 2-DOF and 3-DOF data, with a significant margin on the primary F-Score metric. This is perhaps not surprising, since the OC and 2-DOF VC models are not trained on 3-DOF data but it demonstrates the significant benefit arising from our 3-DOF training approach. The poor performance of 2-DOF VC on 3-DOF VC data shows that such representations may still retain some bias towards the learned shape categories in their canonical pose and cannot reconstruct novel shapes in arbitrary poses as well as the 3-DOF VC model. It can be seen that 3-DOF VC outperforms 2-DOF VC and OC on capturing surface details and concavity of the shapes.

\subsection{Low Image Rendering Variability Negatively Affects Shape Learning}\label{sec:rendering}
It is essential to understand the generalization ability of reconstruction algorithms with respect to changes in lighting, object surface shading,\footnote{We investigate the effect of object surface reflectance properties while the surface texture remains constant} and scene background, since models that use low-level image cues to infer object shape should ideally be robust to such changes. Further, prior works~\cite{mescheder2019occupancy, xu2019disn, choy20163d}, with the exception of~\cite{genre}, primarily train and perform quantitative evaluation of 3D shape reconstruction algorithms on Lambertian shaded objects on white backgrounds. Are there are any limitations to prior empirical approaches? We perform a seen category reconstruction experiment 13 ShapeNet categories using SDFNet trained under three input regimes: image inputs only (SDFNet Img), estimated 2.5D sketch (SDFNet Est), and ground 2.5D (SDFNet Orcl). We generate images under two rendering conditions: (1) Basic with Lambertian shading, uniform, area light sources and white backgrounds (B), (2) varying lighting, reflectance and background (LRBg) (see Fig.~\ref{fig:rendering_effects}). All models are trained under the Basic setting and are then tested on novel objects of the 13 seen categories from both settings.

\begin{table}[t!]
\begin{minipage}[b]{\linewidth}

\begin{center}
\resizebox{\linewidth}{!}{

\begingroup
\setlength{\tabcolsep}{4pt} 
\renewcommand{\arraystretch}{1.2} 

\begin{tabular}{l|ccc|c|ccc|c|ccc|c|ccc|c}
& \multicolumn{4}{c|}{Basic (B) } & \multicolumn{4}{c}{ LR +Background (LRBg) }  \\
\hline
Input Data & CD $\downarrow$ & IoU CD $\uparrow$ & NC $\uparrow$ & FS@1 $\uparrow$  & CD $\downarrow$ & IoU $\uparrow$ & NC $\uparrow$ & FS@1 \\
\hline
SDFNet Img & 0.069 & 0.70 & 0.76 & 0.31 & 0.485 & 0.21 & 0.60 & 0.01 \\   
SDFNet Est   & 0.09 & 0.69 & 0.76 & 0.30 & 0.190 & 0.59 & 0.66 & 0.12  \\     
\hline
SDFNet Orcl  & 0.041 & 0.78 & 0.82 & 0.53 & 0.041 & 0.78 & 0.82 & 0.53 \\
\hline
\end{tabular}

\endgroup

}
\end{center}

    \vspace{-15pt}
    \caption{Reconstruction performance of SDFNet using images (SDFNet Img) and 2.5D predictions (SDFNet Est) is affected by varying lighting, background and reflectance variability. Methods are trained on ShapeNet 3-DOF data with uniform lighting and Lambertian shading and tested on a subset of validation set of seen categories for each rendering variability. SDFNet Orcl, trained on GT 2.5D, included as reference.}
    \label{table:rendering_effects}
\end{minipage}%
\vspace{-15pt}
\end{table}


Our findings show the expected result that the performance of the non-oracle models degrades when tested on data with variable lighting, reflectance and backgrounds (see Tbl.~\ref{table:rendering_effects}). Further, this provides additional evidence for the value of an intermediate 2.5D representation (SDFNet Est), as it exhibits a significantly lower drop in performance than the image only model (SDFNet Img). Our findings suggest that models should therefore be trained on data with high visual variability in order achieve effective generalization to images with high visual variability. Results in Sec.~\ref{sec:method_compare} show good performance for models trained under all sources of variability (LRBg).

\subsection{Cross Dataset Shape Reconstruction}\label{sec:cross_dataset}
\begin{figure}[t!]
\begin{minipage}[t!]{\linewidth}
    \includegraphics[width=\linewidth]{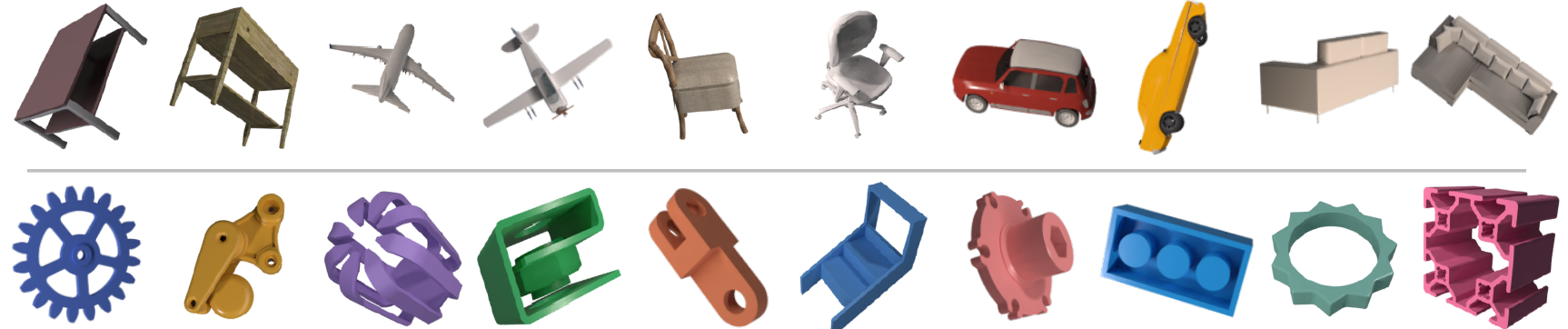}
    \caption{Example objects from the most numerous categories in ShapeNet (top) and  examples of ABC (bottom).}
    \label{fig:domain_gap}
\end{minipage}
\vspace{-10pt}
\end{figure}

In this section we further investigate general 3D shape reconstruction through experiments evaluating cross-dataset generalization. Tatarchenko et al. \cite{tatarchenko2019single} discuss the data leakage issue, which arises when objects in the testing set are similar to objects in the training set. Zhang et al. \cite{genre} propose testing on unseen categories as a more effective test of generalization. In this section, we go beyond testing generalization on novel classes by experimenting with two inherently different datasets: ShapeNetCore.v2 and ABC, illustrated in Fig.~\ref{fig:domain_gap}. For this experiment we train 3-DOF VC SDFNet with ground truth 2.5D sketches as input. For this experiment we decompose the error into the visible and self-occluded object surface components, as shown in Tbl.~\ref{table:domain_gap}.

Our findings show that the performance of the model trained on ABC and evaluated on unseen ShapeNet categories is on par with the performance of the model trained on ShapeNet seen categories and tested on ShapeNet unseen categories for both visible and occluded surfaces. This is a surprising result since we expect the ability to infer occluded surfaces to be biased towards the training data domain. The converse is not true, since 17\% of the generated meshes when training on ShapeNet and testing ABC on are empty. This suggests that the ABC model learns a more robust shape representation that potentially comes from the fact that ABC objects are more diverse in shape. We believe we are the first to show that given access to ground truth 2.5D inputs and 3-DOF data, it is possible to generalize between significantly different shape datasets. Qualitative results are shown in Fig.~\ref{fig:cross_dataset_viz} with further results in Supplement. 

\begin{table}[t!]
\begin{minipage}[t!]{\linewidth}
    
\begin{center}

\resizebox{\linewidth}{!}{

\begingroup
\setlength{\tabcolsep}{4pt} 
\renewcommand{\arraystretch}{1.2} 

\begin{tabular}{l|ccc|c|ccc|c}
Train data $\rightarrow$ &\multicolumn{4}{c|}{ShapeNet}& \multicolumn{4}{c}{ABC} \\
\hline
 Test data $\downarrow$ & CD & IoU & NC & FS@1 & CD & IoU & NC & FS@1\\
\hline
Vis. ShapeNet           & 0.033 & N/A & 0.88 & 0.63 & 0.038 & N/A & 0.87 & 0.57 \\
Occ. ShapeNet         & 0.058 & N/A & 0.82 & 0.38 & 0.062 & N/A & 0.81 & 0.41 \\
Vis. ABC                & 0.643 & N/A & 0.73 & 0.54 & 0.026 & N/A & 0.89 & 0.67 \\
Occ. ABC              & 0.658 & N/A & 0.66 & 0.34 & 0.044 & N/A & 0.82 & 0.55 \\
\hline
ShapeNet                & 0.044 & 0.75 & 0.83 & 0.51 & 0.047 & 0.74 & 0.83 & 0.50 \\
ABC                     & 0.65 & 0.64 & 0.51 & 0.44 & 0.035 & 0.79 & 0.84 & 0.62 \\

\hline
\end{tabular}

\endgroup
}
\end{center}

    \vspace{-10pt}
    \caption{Cross-dataset comparison of generalization to unseen classes of ShapeNet and test samples from ABC and vice versa. All models are trained on 3-DOF viewpoint data with ground truth 2.5D sketches as inputs. The first row parity reports visible (Vis.) and occluded (Occ.) surface performances of each testing dataset; the second row parity reports reconstruction performance overall.}
    \label{table:domain_gap}
\end{minipage}%
\vspace{-10pt}
\end{table}


\begin{figure}[t!]
\centering
\includegraphics[width=\linewidth]{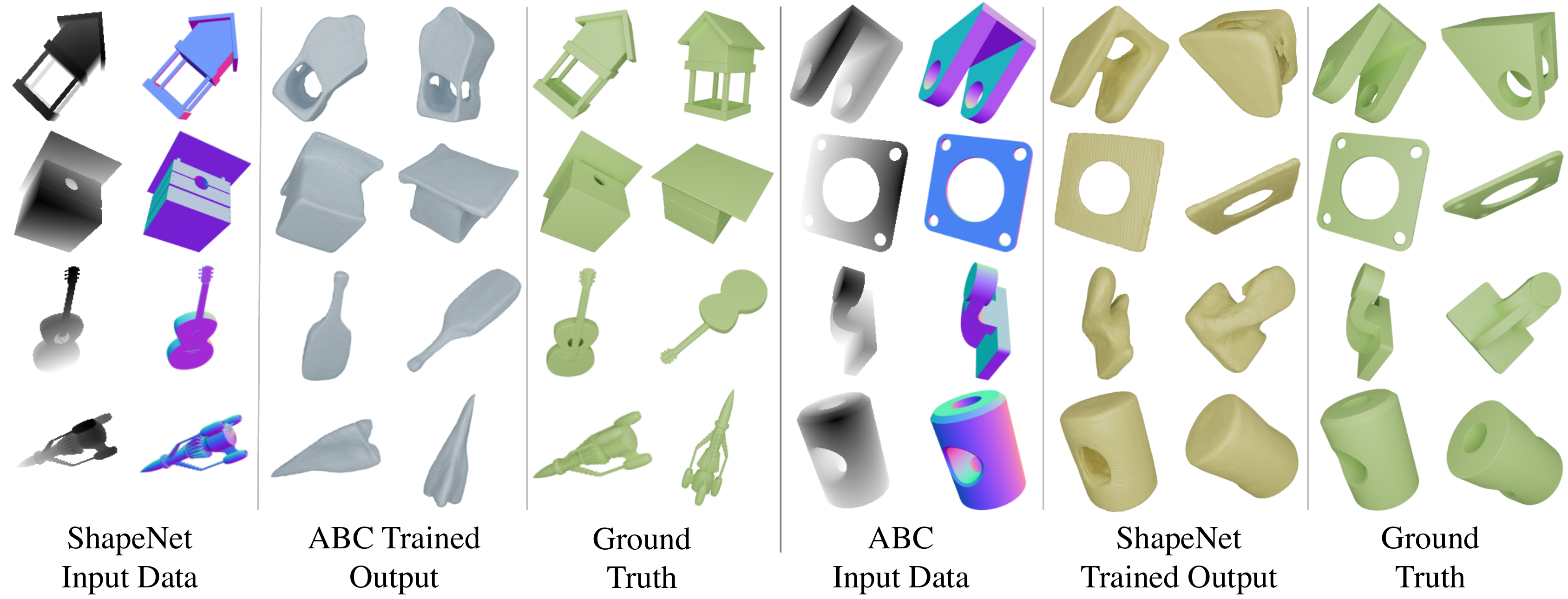}
\vspace{-18pt}
\caption{Cross-dataset generalization performance of a model trained on ABC, tested on unseen classes of ShapeNet (left) and trained on ShapeNet, tested on ABC (right). All models are trained on 3-DOF data with ground truth 2.5D sketches as inputs.}
\label{fig:cross_dataset_viz}
\vspace{-14pt}
\end{figure}


\section{Conclusion}
This paper presents the first comprehensive exploration of generalization of 3D unseen shapes reconstruction, by evaluating on both unseen categories and a completely different dataset. Our solution consists of SDFNet, a novel architecture that combines a 2.5D sketch estimator with a 3D shape regressor that learns a signed distance function. Our findings imply that future approaches to single-view 3D shape reconstruction can benefit significantly from rendering with high visual variability (LRBg) and generating 3-DOF views of models (3-DOF VC representation). In addition, testing on diverse shape datasets is an important future direction for effectively evaluating generalization performance.

\section{Acknowledgement}
We would like to thank Randal Michnovicz for his involvement in discussions and early empirical investigations. This work was supported by NSF Award 1936970 and NIH R01-MH114999.
{\small
\bibliographystyle{ieee_fullname}
\bibliography{egbib}

\begin{thebibliography}{10}\itemsep=-1pt

\bibitem{banani2020novel}
Mohamed~El Banani, Jason~J Corso, and David~F Fouhey.
\newblock Novel object viewpoint estimation through reconstruction alignment.
\newblock In {\em Proceedings of the IEEE/CVF Conference on Computer Vision and
  Pattern Recognition}, pages 3113--3122, 2020.

\bibitem{barrow1978recovering}
Harry Barrow, J Tenenbaum, A Hanson, and E Riseman.
\newblock Recovering intrinsic scene characteristics.
\newblock {\em Comput. Vis. Syst}, 2(3-26):2, 1978.

\bibitem{baslamisli2018cnn}
Anil~S Baslamisli, Hoang-An Le, and Theo Gevers.
\newblock Cnn based learning using reflection and retinex models for intrinsic
  image decomposition.
\newblock In {\em Proceedings of the IEEE Conference on Computer Vision and
  Pattern Recognition}, pages 6674--6683, 2018.

\bibitem{bautista2021generalization}
Miguel~Angel Bautista, Walter Talbott, Shuangfei Zhai, Nitish Srivastava, and
  Joshua~M Susskind.
\newblock On the generalization of learning-based 3d reconstruction.
\newblock In {\em Proceedings of the IEEE/CVF Winter Conference on Applications
  of Computer Vision}, pages 2180--2189, 2021.

\bibitem{blender}
{Blender Online Community}.
\newblock {\em Blender - a 3D modelling and rendering package}.
\newblock Blender Foundation, Blender Institute, Amsterdam.

\bibitem{chabra2020deepLS}
Rohan Chabra, Jan~E. Lenssen, Eddy Ilg, Tanner Schmidt, Julian Straub, Steven
  Lovegrove, and Richard Newcombe.
\newblock Deep local shapes: Learning local sdf priors for detailed 3d
  reconstruction.
\newblock In {\em Computer Vision -- ECCV 2020}, pages 608--625. Springer
  International Publishing, 2020.

\bibitem{chang2015shapenet}
Angel~X Chang, Thomas Funkhouser, Leonidas Guibas, Pat Hanrahan, Qixing Huang,
  Zimo Li, Silvio Savarese, Manolis Savva, Shuran Song, Hao Su, et~al.
\newblock Shapenet: An information-rich 3d model repository.
\newblock {\em arXiv preprint arXiv:1512.03012}, 2015.

\bibitem{chen2016single}
Weifeng Chen, Zhao Fu, Dawei Yang, and Jia Deng.
\newblock Single-image depth perception in the wild.
\newblock In {\em Advances in neural information processing systems}, pages
  730--738, 2016.

\bibitem{chen2019learning}
Zhiqin Chen and Hao Zhang.
\newblock Learning implicit fields for generative shape modeling.
\newblock In {\em Proceedings of the IEEE Conference on Computer Vision and
  Pattern Recognition}, pages 5939--5948, 2019.

\bibitem{choy20163d}
Christopher~B Choy, Danfei Xu, JunYoung Gwak, Kevin Chen, and Silvio Savarese.
\newblock 3d-r2n2: A unified approach for single and multi-view 3d object
  reconstruction.
\newblock In {\em European conference on computer vision}, pages 628--644.
  Springer, 2016.

\bibitem{de2017modulating}
Harm De~Vries, Florian Strub, J{\'e}r{\'e}mie Mary, Hugo Larochelle, Olivier
  Pietquin, and Aaron~C Courville.
\newblock Modulating early visual processing by language.
\newblock In {\em Advances in Neural Information Processing Systems}, pages
  6594--6604, 2017.

\bibitem{facil2019cam}
Jose~M Facil, Benjamin Ummenhofer, Huizhong Zhou, Luis Montesano, Thomas Brox,
  and Javier Civera.
\newblock Cam-convs: camera-aware multi-scale convolutions for single-view
  depth.
\newblock In {\em Proceedings of the IEEE conference on computer vision and
  pattern recognition}, pages 11826--11835, 2019.

\bibitem{Firman_2016_CVPR}
Michael Firman, Oisin Mac~Aodha, Simon Julier, and Gabriel~J. Brostow.
\newblock Structured prediction of unobserved voxels from a single depth image.
\newblock In {\em The IEEE Conference on Computer Vision and Pattern
  Recognition (CVPR)}, June 2016.

\bibitem{Giancola2019LeveragingSC}
Silvio Giancola, Jesus Zarzar, and Bernard Ghanem.
\newblock Leveraging shape completion for 3d siamese tracking.
\newblock {\em 2019 IEEE/CVF Conference on Computer Vision and Pattern
  Recognition (CVPR)}, pages 1359--1368, 2019.

\bibitem{groueix2018atlasnet}
Thibault Groueix, Matthew Fisher, Vladimir~G Kim, Bryan~C Russell, and Mathieu
  Aubry.
\newblock Atlasnet: A papier-m$\backslash$\^{} ach$\backslash$'e approach to
  learning 3d surface generation.
\newblock {\em arXiv preprint arXiv:1802.05384}, 2018.

\bibitem{henderson2019learning}
Paul Henderson and Vittorio Ferrari.
\newblock Learning single-image 3d reconstruction by generative modelling of
  shape, pose and shading.
\newblock {\em International Journal of Computer Vision}, pages 1--20, 2019.

\bibitem{huang2019framenet}
Jingwei Huang, Yichao Zhou, Thomas Funkhouser, and Leonidas~J Guibas.
\newblock Framenet: Learning local canonical frames of 3d surfaces from a
  single rgb image.
\newblock In {\em Proceedings of the IEEE International Conference on Computer
  Vision}, pages 8638--8647, 2019.

\bibitem{ikeuchi1981numerical}
Katsushi Ikeuchi and Berthold~KP Horn.
\newblock Numerical shape from shading and occluding boundaries.
\newblock {\em Artificial intelligence}, 17(1-3):141--184, 1981.

\bibitem{jack2018learning}
Dominic Jack, Jhony~K Pontes, Sridha Sridharan, Clinton Fookes, Sareh Shirazi,
  Frederic Maire, and Anders Eriksson.
\newblock Learning free-form deformations for 3d object reconstruction.
\newblock In {\em Asian Conference on Computer Vision}, pages 317--333.
  Springer, 2018.

\bibitem{kingma2014adam}
Diederik~P Kingma and Jimmy Ba.
\newblock Adam: A method for stochastic optimization.
\newblock {\em arXiv preprint arXiv:1412.6980}, 2014.

\bibitem{kleineberg2020adversarial}
Marian Kleineberg, Matthias Fey, and Frank Weichert.
\newblock Adversarial generation of continuous implicit shape representations.
\newblock {\em arXiv preprint arXiv:2002.00349}, 2020.

\bibitem{koch2019abc}
Sebastian Koch, Albert Matveev, Zhongshi Jiang, Francis Williams, Alexey
  Artemov, Evgeny Burnaev, Marc Alexa, Denis Zorin, and Daniele Panozzo.
\newblock Abc: A big cad model dataset for geometric deep learning.
\newblock In {\em Proceedings of the IEEE Conference on Computer Vision and
  Pattern Recognition}, pages 9601--9611, 2019.

\bibitem{kuznietsov2017semi}
Yevhen Kuznietsov, Jorg Stuckler, and Bastian Leibe.
\newblock Semi-supervised deep learning for monocular depth map prediction.
\newblock In {\em Proceedings of the IEEE conference on computer vision and
  pattern recognition}, pages 6647--6655, 2017.

\bibitem{Li2018CGIntrinsicsBI}
Zhengqi Li and Noah Snavely.
\newblock Cgintrinsics: Better intrinsic image decomposition through
  physically-based rendering.
\newblock In {\em ECCV}, 2018.

\bibitem{Li_2018_CVPR}
Zhengqi Li and Noah Snavely.
\newblock Megadepth: Learning single-view depth prediction from internet
  photos.
\newblock In {\em The IEEE Conference on Computer Vision and Pattern
  Recognition (CVPR)}, June 2018.

\bibitem{li2018learning}
Zhengqin Li, Zexiang Xu, Ravi Ramamoorthi, Kalyan Sunkavalli, and Manmohan
  Chandraker.
\newblock Learning to reconstruct shape and spatially-varying reflectance from
  a single image.
\newblock {\em ACM Transactions on Graphics (TOG)}, 37(6):1--11, 2018.

\bibitem{Liu_2019_ICCV}
Shichen Liu, Tianye Li, Weikai Chen, and Hao Li.
\newblock Soft rasterizer: A differentiable renderer for image-based 3d
  reasoning.
\newblock In {\em The IEEE International Conference on Computer Vision (ICCV)},
  October 2019.

\bibitem{liu2019learning}
Shichen Liu, Shunsuke Saito, Weikai Chen, and Hao Li.
\newblock Learning to infer implicit surfaces without 3d supervision.
\newblock In {\em Advances in Neural Information Processing Systems}, pages
  8293--8304, 2019.

\bibitem{lorensen1987marching}
William~E Lorensen and Harvey~E Cline.
\newblock Marching cubes: A high resolution 3d surface construction algorithm.
\newblock {\em ACM siggraph computer graphics}, 21(4):163--169, 1987.

\bibitem{mescheder2019occupancy}
Lars Mescheder, Michael Oechsle, Michael Niemeyer, Sebastian Nowozin, and
  Andreas Geiger.
\newblock Occupancy networks: Learning 3d reconstruction in function space.
\newblock In {\em Proceedings of the IEEE Conference on Computer Vision and
  Pattern Recognition}, pages 4460--4470, 2019.

\bibitem{mildenhall2020nerf}
Ben Mildenhall, Pratul~P Srinivasan, Matthew Tancik, Jonathan~T Barron, Ravi
  Ramamoorthi, and Ren Ng.
\newblock Nerf: Representing scenes as neural radiance fields for view
  synthesis.
\newblock In {\em European Conference on Computer Vision}, pages 405--421.
  Springer, 2020.

\bibitem{park2019deepsdf}
Jeong~Joon Park, Peter Florence, Julian Straub, Richard Newcombe, and Steven
  Lovegrove.
\newblock Deepsdf: Learning continuous signed distance functions for shape
  representation.
\newblock {\em arXiv preprint arXiv:1901.05103}, 2019.

\bibitem{Qi_2018_CVPR}
Xiaojuan Qi, Renjie Liao, Zhengzhe Liu, Raquel Urtasun, and Jiaya Jia.
\newblock Geonet: Geometric neural network for joint depth and surface normal
  estimation.
\newblock In {\em The IEEE Conference on Computer Vision and Pattern
  Recognition (CVPR)}, June 2018.

\bibitem{Qiu_2019_CVPR}
Jiaxiong Qiu, Zhaopeng Cui, Yinda Zhang, Xingdi Zhang, Shuaicheng Liu, Bing
  Zeng, and Marc Pollefeys.
\newblock Deeplidar: Deep surface normal guided depth prediction for outdoor
  scene from sparse lidar data and single color image.
\newblock In {\em The IEEE Conference on Computer Vision and Pattern
  Recognition (CVPR)}, June 2019.

\bibitem{Rock_2015_CVPR}
Jason Rock, Tanmay Gupta, Justin Thorsen, JunYoung Gwak, Daeyun Shin, and Derek
  Hoiem.
\newblock Completing 3d object shape from one depth image.
\newblock In {\em The IEEE Conference on Computer Vision and Pattern
  Recognition (CVPR)}, June 2015.

\bibitem{ronneberger2015unet}
Olaf Ronneberger, Philipp Fischer, and Thomas Brox.
\newblock U-net: Convolutional networks for biomedical image segmentation.
\newblock In {\em International Conference on Medical image computing and
  computer-assisted intervention}, pages 234--241. Springer, 2015.

\bibitem{shin2018pixels}
Daeyun Shin, Charless~C Fowlkes, and Derek Hoiem.
\newblock Pixels, voxels, and views: A study of shape representations for
  single view 3d object shape prediction.
\newblock In {\em Proceedings of the IEEE Conference on Computer Vision and
  Pattern Recognition}, pages 3061--3069, 2018.

\bibitem{smith19a}
Edward Smith, Scott Fujimoto, Adriana Romero, and David Meger.
\newblock {GEOM}etrics: Exploiting geometric structure for graph-encoded
  objects.
\newblock In Kamalika Chaudhuri and Ruslan Salakhutdinov, editors, {\em
  Proceedings of the 36th International Conference on Machine Learning},
  volume~97 of {\em Proceedings of Machine Learning Research}, pages
  5866--5876, Long Beach, California, USA, 09--15 Jun 2019. PMLR.

\bibitem{Song_2017_CVPR}
Shuran Song, Fisher Yu, Andy Zeng, Angel~X. Chang, Manolis Savva, and Thomas
  Funkhouser.
\newblock Semantic scene completion from a single depth image.
\newblock In {\em The IEEE Conference on Computer Vision and Pattern
  Recognition (CVPR)}, July 2017.

\bibitem{stutz2018learning}
David Stutz and Andreas Geiger.
\newblock Learning 3d shape completion under weak supervision.
\newblock {\em International Journal of Computer Vision}, pages 1--20, 2018.

\bibitem{su2015multi}
Hang Su, Subhransu Maji, Evangelos Kalogerakis, and Erik Learned-Miller.
\newblock Multi-view convolutional neural networks for 3d shape recognition.
\newblock In {\em Proceedings of the IEEE international conference on computer
  vision}, pages 945--953, 2015.

\bibitem{tappen2003recovering}
Marshall~F Tappen, William~T Freeman, and Edward~H Adelson.
\newblock Recovering intrinsic images from a single image.
\newblock In {\em Advances in neural information processing systems}, pages
  1367--1374, 2003.

\bibitem{tatarchenko2016multi}
Maxim Tatarchenko, Alexey Dosovitskiy, and Thomas Brox.
\newblock Multi-view 3d models from single images with a convolutional network.
\newblock In {\em European Conference on Computer Vision}, pages 322--337.
  Springer, 2016.

\bibitem{tatarchenko2019single}
Maxim Tatarchenko, Stephan~R Richter, Ren{\'e} Ranftl, Zhuwen Li, Vladlen
  Koltun, and Thomas Brox.
\newblock What do single-view 3d reconstruction networks learn?
\newblock In {\em Proceedings of the IEEE Conference on Computer Vision and
  Pattern Recognition}, pages 3405--3414, 2019.

\bibitem{wanggsir}
Jianren Wang and Zhaoyuan Fang.
\newblock Gsir: Generalizable 3d shape interpretation and reconstruction.
\newblock In {\em Proceedings of the European conference on computer vision
  (ECCV)}, 2020.

\bibitem{wang2018pixel2mesh}
Nanyang Wang, Yinda Zhang, Zhuwen Li, Yanwei Fu, Wei Liu, and Yu-Gang Jiang.
\newblock Pixel2mesh: Generating 3d mesh models from single rgb images.
\newblock In {\em Proceedings of the European Conference on Computer Vision
  (ECCV)}, pages 52--67, 2018.

\bibitem{wernerTSDF}
Diana Werner, Ayoub Al-Hamadi, and Philipp Werner.
\newblock Truncated signed distance function: Experiments on voxel size.
\newblock In Aur{\'e}lio Campilho and Mohamed Kamel, editors, {\em Image
  Analysis and Recognition}, pages 357--364, Cham, 2014. Springer International
  Publishing.

\bibitem{wu2017marrnet}
Jiajun Wu, Yifan Wang, Tianfan Xue, Xingyuan Sun, Bill Freeman, and Josh
  Tenenbaum.
\newblock Marrnet: 3d shape reconstruction via 2.5 d sketches.
\newblock In {\em Advances in neural information processing systems}, pages
  540--550, 2017.

\bibitem{wu2015modelnet}
Zhirong Wu, Shuran Song, Aditya Khosla, Fisher Yu, Linguang Zhang, Xiaoou Tang,
  and Jianxiong Xiao.
\newblock 3d shapenets: A deep representation for volumetric shapes.
\newblock In {\em Proceedings of the IEEE conference on computer vision and
  pattern recognition}, pages 1912--1920, 2015.

\bibitem{wu20153d}
Zhirong Wu, Shuran Song, Aditya Khosla, Fisher Yu, Linguang Zhang, Xiaoou Tang,
  and Jianxiong Xiao.
\newblock 3d shapenets: A deep representation for volumetric shapes.
\newblock In {\em Proceedings of the IEEE conference on computer vision and
  pattern recognition}, pages 1912--1920, 2015.

\bibitem{genre-git}
Z.~Zhang et~al. X.~Zhang.
\newblock https://github.com/xiumingzhang/GenRe-ShapeHD.

\bibitem{xiao2010sun}
Jianxiong Xiao, James Hays, Krista~A Ehinger, Aude Oliva, and Antonio Torralba.
\newblock Sun database: Large-scale scene recognition from abbey to zoo.
\newblock In {\em 2010 IEEE Computer Society Conference on Computer Vision and
  Pattern Recognition}, pages 3485--3492. IEEE, 2010.

\bibitem{xu2019disn}
Qiangeng Xu, Weiyue Wang, Duygu Ceylan, Radomir Mech, and Ulrich Neumann.
\newblock Disn: Deep implicit surface network for high-quality single-view 3d
  reconstruction.
\newblock 2019.

\bibitem{Yang_2017_ICCV}
Bo Yang, Hongkai Wen, Sen Wang, Ronald Clark, Andrew Markham, and Niki Trigoni.
\newblock 3d object reconstruction from a single depth view with adversarial
  learning.
\newblock In {\em The IEEE International Conference on Computer Vision (ICCV)
  Workshops}, Oct 2017.

\bibitem{Zeng2019DeepSN}
Jin Zeng, Yanfeng Tong, Yunmu Huang, Qiong Yan, Wenxiu Sun, Jing Chen, and
  Yongtian Wang.
\newblock Deep surface normal estimation with hierarchical rgb-d fusion.
\newblock {\em 2019 IEEE/CVF Conference on Computer Vision and Pattern
  Recognition (CVPR)}, pages 6146--6155, 2019.

\bibitem{genre}
Xiuming Zhang, Zhoutong Zhang, Chengkai Zhang, Joshua~B Tenenbaum, William~T
  Freeman, and Jiajun Wu.
\newblock {Learning to Reconstruct Shapes from Unseen Classes}.
\newblock In {\em Advances in Neural Information Processing Systems (NeurIPS)},
  2018.

\end{thebibliography}
}
\clearpage
\appendix
\label{sec:append}
\section{Results for Cross Dataset Shape Reconstruction}
\label{sec:abc_shapenet_results}

\begin{figure*}[!ht]
\centering
\includegraphics[width=0.5\linewidth]{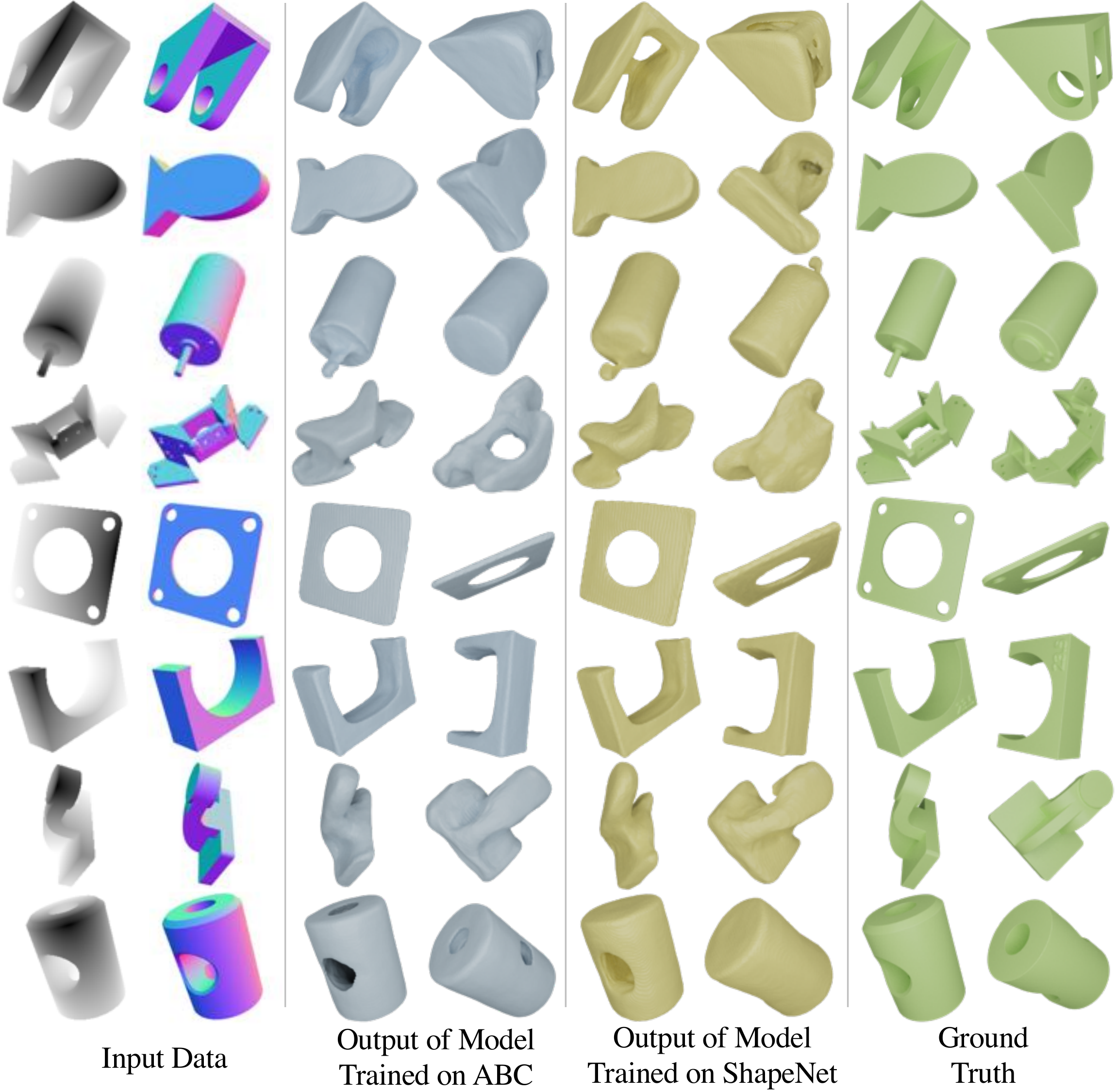}
\caption{Performance on test samples of ABC of models trained on ABC (second column) and ShapeNet (third column) respectively. All models are trained on 3-DOF data with ground truth 2.5D sketches as inputs.}
\vspace{-10pt}
\label{fig:abc_and_shapenet_on_abc}
\end{figure*}

Figure~\ref{fig:abc_and_shapenet_on_abc} contains qualitative results for SDFNet trained on ground truth 2.5D sketches of ABC and ShapeNet (seen classes) and tested on the ABC test set. The outputs suggest that the model trained on ABC has a better ability to capture shape detail and non-convexity e.g. hole in the fourth and the last row, spindle protruding from the cylinder the third row.
\section{Comparison with GenRe}
\label{sec:genre_comparison}
In this section we describe the steps we took to perform a comparison with GenRe on the complete ShapeNetCore.v2 dataset. GenRe~\cite{genre} is a three stage method consisting of depth estimation, spherical inpainting, and voxel refinement, that are trained sequentially one after the other. Each stage requires its own set of ground truth data for training, which the authors have released for the three largest classes of ShapeNet. For testing, the authors have released around 30 objects for each of the 9 unseen classes. In order to run GenRe~\cite{genre} on our training and testing split of 13 seen and 42 unseen classes, we re-implemented the ground truth data generation pipeline for GenRe (referred to as GenRe GT Pipeline for the rest of this section). To generate RGB images and ground truth depth images, we adapted our Blender-based image rendering pipeline. To generate the full spherical maps, we partially adapted code from the authors' release, in addition to writing new code to complete the procedure. To produce the voxel grids used during training, we employed code released by DISN \cite{xu2019disn} to extract $128^3$ grids of signed distance fields for each mesh, which are then rotated, re-sampled and truncated to generate a voxel grid for each object view. The original GenRe voxel ground truth data is generated from inverted truncated unsigned distance (TuDF) fields~\cite{wernerTSDF}. In our GenRe experiments Section~4.3, we use a truncation value of $\frac{3}{128}$.

We validated our GenRe GT Pipeline by recreating the training and testing data released by the authors, and comparing the performance of GenRe trained on the released data \cite{genre-git} and data generated by our pipeline. Note that the GenRe authors focus on using CD for evaluation and report the best average CD after sweeping the isosurface threshold for mesh generation using marching cubes \cite{lorensen1987marching} prior to evaluation. In contrast, for CD we sample 100K surface points from the model output mesh and 300K surface points for the ground truth mesh, and use a fixed threshold of 0.25 to generate meshes from the model output. We used a fixed threshold to avoid biasing the performance based on the testing data. The ground truth meshes used to evaluate GenRe are obtained by running Marching Cubes on the TuDF grids using a threshold of 0.5. The model and ground truth meshes are normalized to fit inside a unit cube prior to computing metrics, as in our other experiments.

In Tbl.~\ref{table:validate_pipeline_seen} and Tbl.~\ref{table:validate_pipeline_unseen} we present the outcome of training and testing GenRe on the data provided by the authors, compared with training and testing GenRe on data from our GenRe GT Pipeline. Training on data from our GenRe GT Pipeline results in lower performance than originally reported, resulting in a mean 0.168 CD on unseen classes compared to the reported 0.106 in the paper (Table 1 \cite{genre}). This result is is shown in the first column of Tbl.~\ref{table:validate_pipeline_unseen}. This is potentially since our evaluation procedure is different from the one originally used. We do not sweep threshold values for isosurfaces, we scale the meshes to fit in a unit cube, and sample 100K+ points on the object surface in comparison to 1024 in the evaluation done in GenRe \cite{genre}. The last columns of Tbl.~\ref{table:validate_pipeline_seen} and Tbl.~\ref{table:validate_pipeline_unseen} show comparable the testing performance for GenRe trained using our GenRe GT Pipeline and GenRe trained on the released data. The insignificant difference in performance between training GenRE on the released data and using our GenRe GT Pipeline demonstrates that our implementation is correct.

\begin{table*}[!ht]
    \centering
    \begin{tabular}{c|c|c|c|c}
     & \makecell{CD Reported in \\ Table 1 \cite{genre}}  &  \makecell{Our Evaluation \\ w/o Scaling} & \makecell{Our Evaluation \\ w/ Scaling} & \makecell{Our Implementation \\ w/ Scaling} \\
    \hline
    car & N/A & 0.077 &  0.088 & 0.119 \\

    airplane & N/A & 0.115 & 0.147 & 0.147 \\

    chair & N/A & 0.105 & 0.130 & 0.132 \\
    \hline
    avg & 0.064 & 0.093 & 0.122 & 0.133\\
    \end{tabular}
    \vspace{5pt}
    \caption{Comparison of GenRe performance trained on data from our data generation pipeline with GenRe trained on the data released with the paper on seen categories. Our Evaluation w/o Scaling indicates evaluation without scaling the meshes up to fit in a unit cube, w/ Scaling indicates that meshes have been scaled up to fit inside a unit cube. The last column is trained using our GT generating pipeline and our evaluation code.}
    \label{table:validate_pipeline_unseen}
\end{table*}

\begin{table*}[!ht]
    \centering
    \begin{tabular}{c|c|c|c|c}
     & \makecell{CD Reported in \\ Table 1 \cite{genre}}  &  \makecell{Our Evaluation \\ w/o Scaling} & \makecell{Our Evaluation \\ w/ Scaling} & \makecell{Our Implementation \\ w/ Scaling} \\
    \hline
    bench          & 0.089 & 0.114 & 0.132 & 0.135 \\
    display        & 0.092 & 0.108 & 0.222 & 0.216 \\
    lamp           & 0.124 & 0.177 & 0.225 & 0.196 \\
    loudspeaker    & 0.115 & 0.133 & 0.157 & 0.186\\
    rifle          & 0.112 & 0.149 & 0.183 & 0.157\\
    sofa           & 0.082 & 0.097 & 0.113 & 0.126\\
    table          & 0.096 & 0.129 & 0.151 & 0.169\\
    telephone      & 0.107 & 0.116 & 0.134 & 0.146\\
    vessel         & 0.092 & 0.113 & 0.164 & 0.181\\
    \hline
    avg & 0.106 & 0.137 & 0.165 & 0.168\\
    \end{tabular}
    \vspace{5pt}
    \caption{Comparison of GenRe performance trained on data from our data generation pipeline with GenRe trained on the data released with the paper on unseen categories. Our Evaluation w/o Scaling indicates evaluation without scaling the meshes up to fit in a unit cube, w/ Scaling indicates that meshes have been scaled up to fit inside a unit cube. The last column is trained using our GT generating pipeline and our evaluation code. }
    \label{table:validate_pipeline_seen}
\end{table*}

\section{SDF Point Sampling Strategy Details}
\label{sec:sampling}


We first rescale the mesh to fit inside a unit cube and set the mesh origin to the center of this cube, then we sample 50\% of the training points within a distance of 0.03 to the surface, 80\% within a distance of 0.1 and 20\% randomly in a cube volume of size 1.2. Note that since we are training with viewer-centered coordinate system where the pose of the testing data is unknown, it is important that training signals come from 3D points sampled from a volume of sufficient size. This ensures that during mesh generation at inference, the algorithm does not need to extrapolate to points outside of the training range.


\section{Further Discussion of Metrics}
\label{sec:metrics_further}
\subsection{Issues with Current Metrics}

Metrics for measuring the distance between two 3D shapes play an important role in shape reconstruction. While there are a variety of widely-used metrics, prior work~\cite{tatarchenko2019single,shin2018pixels} has identified significant disadvantages with several standard metrics. IoU has been used extensively, and has the advantage of being straightforward to evaluate since it does not require correspondence between surfaces. However, while IoU is effective in capturing shape similarity at a coarse level, it is difficult to capture fine-grained shape details using IoU, since it is dominated by the interior volume of the shape rather than the surface details~\cite{tatarchenko2019single}. Figure 5 in the main text illustrates some of the issues that can arise in using shape metrics. On the right, the drum is progressively simplified from right to left. The IoU score exhibits very little change, reflecting its poor performance in capturing fine-grained details. In contrast, the F-Score (FS) at 1\% of the side-length of reconstructed volume, in this case a unit side length bound cube, shows good sensitivity to the loss of fine-grained details. On the left of Fig.~4, the bowl is progressively thickened from right to left, while the shape of its surfaces remains largely constant. This example points out two issues. The first is the difficulty of interpreting IoU values for thin objects. An IoU of 0.15 would generally be thought to denote very poor agreement, while in fact the leftmost bowl is a fairly good approximation of the shape of the source bowl. In contrast, normal consistency (NC) is very sensitive to fine-grained shape details but fails to capture volumetric changes, as in this example NC exhibits almost no change despite the progressive thickening. While there is no ideal shape metric, we follow~\cite{tatarchenko2019single} in adopting FS@$d$ as the primary shape metric in this work.



\begin{figure*}[!ht]
\centering
\includegraphics[width=0.8\linewidth]{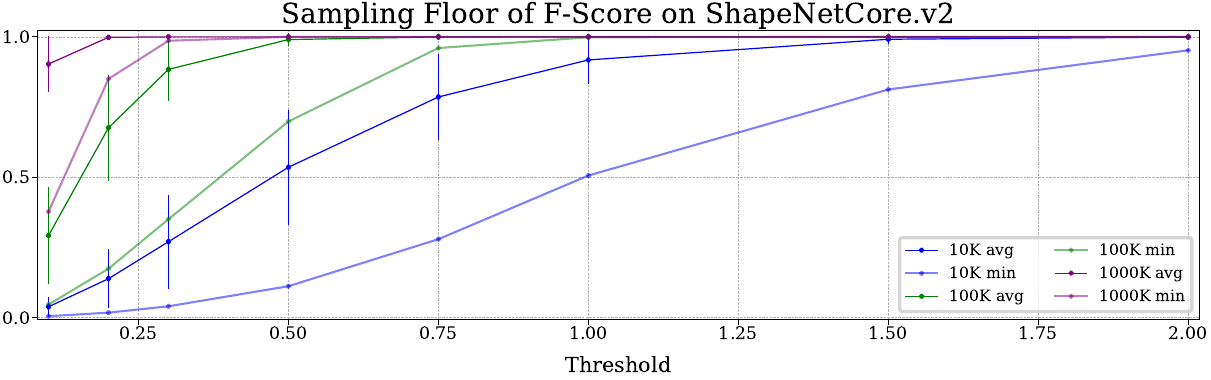}
\caption{An illustration of the Sampling floor (maximum measurable accuracy in comparing two meshes as a function of the number of point samples) on ShapeNetCore.v2 using F-Score at thresholds $d\leq 2\%$. Curves with error bars give the average sampling floor for different numbers of samples from 10K to 1M. Curves without error bars denote the worst-case meshes. Higher is better for FS}
\vspace{-10pt}
\label{fig:fscore}
\end{figure*}

\begin{figure*}[!ht]
\centering
\includegraphics[width=0.8\linewidth]{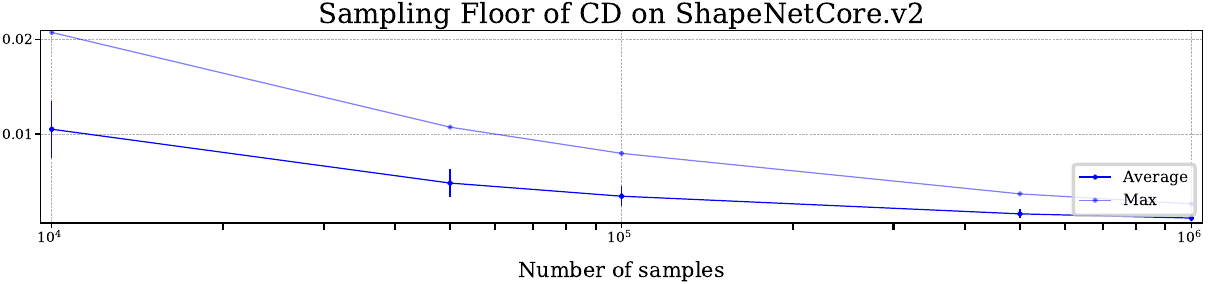}
\caption{An illustration of the Sampling floor (maximum measurable accuracy in comparing two meshes as a function of the number of point samples) on ShapeNetCore.v2 using Chamfer Distance (CD). Curves with error bars give the average sampling floor for different numbers of samples from 10K to 1M. Curves without error bars denote the worst-case meshes. Lower is better for CD.}
\vspace{-10pt}
\label{fig:cd}
\end{figure*}

\begin{figure*}[!ht]
\centering
\includegraphics[width=0.8\linewidth]{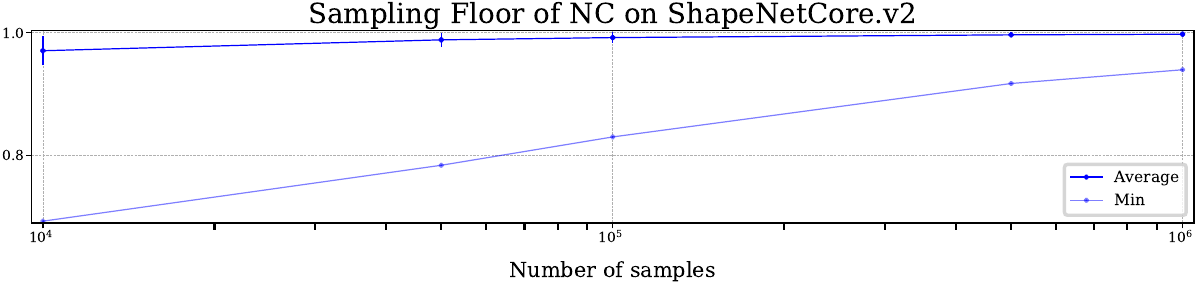}
\caption{An illustration of the Sampling floor (maximum measurable accuracy in comparing two meshes as a function of the number of point samples) on ShapeNetCore.v2 using Normal Consistency (NC). Curves with error bars give the average sampling floor for different numbers of samples from 10K to 1M. Curves without error bars denote the worst-case meshes. Higher is better for NC.}
\vspace{-10pt}
\label{fig:nc}
\end{figure*}

The sampling floor issue is illustrated in Figs. ~\ref{fig:fscore},~\ref{fig:cd} and ~\ref{fig:nc}. To generate the curves in each figure, we take each object in ShapeNetCore.v2 and treat it as both the source and target object in computing the shape metrics. For example, with F-Score, we randomly sample the indicated number of surface points twice, to obtain both source and target point clouds, and then compare the point clouds under the FS@$d$ shape metric for different choices of $d$ (thresholds, along x axis). The average curves plot the average accuracy score (y axis, with error bars) for each choice of $d$. The minimum curves denote the minimum FS for the single worst-case mesh at each threshold. Since the source and target objects are identical meshes, the metric should always be 1, denoting perfect similarity. We can see that for 10K samples the average FS@1 score is around 0.8, which is an upper bound on the ability to measure the reconstruction accuracy under this evaluation approach. A practical constraint on the use of a large number of samples is the time complexity of Nearest Neighbor matching ($O(N\text{log}N)$). Evaluation times of around 2 hours are required for 100K points on 10K meshes on a Titan X GPU and 12 CPU threads. For 1M point samples, evaluation would take approximately 2 days, which is twice as long as the time required to train the model.  Note that the value of the sampling floor will depend upon the choice of both the dataset and shape metric.


The sampling floor graphs (Figs.~\ref{fig:fscore},~\ref{fig:cd},~\ref{fig:nc}) can be used to select an appropriate number of points for all metrics (CD, NC and FS) and a corresponding threshold $d$ for F-Score. For example,~\cite{tatarchenko2019single} has suggested using the F-Score metric at a threshold $\leq 2\%$ of the side length of the reconstructed volume. We can see that in order to achieve an average sampling floor of 0.9 or higher for ShapeNetCore55, 100K samples would be needed for FS@0.5 or higher, while 10K samples are acceptable for FS@1.5 or higher. Note that the significant gap between the worst and average cases in Figs.~\ref{fig:fscore},~\ref{fig:cd},~\ref{fig:nc} implies that for a subset of the meshes, the sampling error is significantly worse than in the average case. This suggests that it may be beneficial to compute and report the sampling floor along with their performance evaluation on any novel datasets in order to provide appropriate context.

\subsection{Implementation of Metrics}
In this work, we provide implementations for IoU, Chamfer distance (CD), normal consistency (NC) and F-score@d (FS). For IoU we sample 3D points and compute $\text{IoU}=\vert O_1 \cap O_2\vert/\vert O_1 \cup O_2\vert$ where $O_1$ and $O_2$ are occupancy values obtained from checking whether the sampled points are inside or outside the meshes. For testing purposes, we sample points densely near the surface of the ground truth mesh with the same density as training. Note that this way of computing IoU will more strictly penalize errors made near the mesh surface. This therefore captures fine-grained shape details better than IoU computed after voxelization or using grid occupancies. IoU computed from grid occupancies requires high resolution to precisely capture thin parts of shapes which can be computationally expensive. Although we sample points densely closer to the mesh surface, to accurately measure IoU, it is important to also sample enough points uniformly in the cube volume.  

For CD, NC and FS, we first sample 300K points and 100K points on the surface of predicted mesh ($S_1$) and ground-truth mesh ($S_2$) respectively. The metrics are computed as follows
$$CD(S_1,S_2)=\frac{1}{\vert S_1 \vert}\sum_{x\in S_1}\min_{y\in S_2}\Vert x-y\Vert_2+\frac{1}{\vert S_2\vert}\sum_{y\in S_2}\min_{x\in S_1}\Vert x-y\Vert_2,$$
\begin{align*}
    NC(S_1,S_2)&=\frac{1}{2\vert S_1 \vert}\sum_{x\in S_1}\min_{y\in S_2}\vert\langle\vec{n}_x,\vec{n}_y\rangle\vert\\
    &+\frac{1}{2\vert S_2\vert}\sum_{y\in S_2}\min_{x\in S_1}\vert\langle\vec{n}_x,\vec{n}_y\rangle\vert,
\end{align*}
where $\vec{n}_x$ denotes the normal vector at point $x$, and 
$$FS@d(S_1,S_2)=\frac{2\cdot Precision@d \cdot Recall@d }{Precision@d+Recall@d},$$ where $Precision@d$ measures the portion of points from the predicted mesh that lie within a threshold $d$ to the points from the ground truth mesh, and $Recall@d$ indicates the portion of points from the ground truth mesh that lie within a threshold $d$ to the points from the predicted mesh.

\section{Further Data Generation Details}
\label{sec:further_data_gen}

\emph{Object Origin:} Current single-view object shape reconstruction algorithms are not robust to changes in object translation, with training generally done with the object at the center of the image. This requires careful consideration of the placement of the object origin when performing rotation. Object meshes in ShapeNet have a predetermined origin. The GenRe algorithm is implemented for ShapeNet so that objects rotate around this object origin for VC training. For experiments with GenRe, we kept to this original design decision and rendered the objects after rotating them about the predetermined origin. For SDFNet, we rotate the object about the center of its bounding box. As a result of this distinction, GenRe and SDFNet are trained and tested on two distinct sets of object renderings, consisting of the same objects, with the same pose variability, rendered under the same lighting and reflectance variability settings, with the only difference being the object origin. This difference can be seen in Figure~3 in the main text.

\emph{Pose Variability During 2-DOF VC and 3-DOF VC training:} For 2-DOF training,  we render views in the range of $[-50,50]$ for elevation and $[0,360]$ for azimuth. For 3-DOF VC, in order to include tilt, and achieve high variability in object pose, we initially apply a random pose to the object, and then generate 25 views using the same procedure and parameters as for the 2-DOF case. 

\emph{Camera Parameters} For all generated data, the camera distance is 2.2 from the the origin, where the object is placed. The focal length of the camera is 50mm with a 32mm sensor size. All images are rendered with a 1:1 aspect ratio and at a resolution of $256 \times 256$ pixels.

\emph{Background} While it is possible is to use image based lighting techniques such as environment mapping to generate variability in backgrounds and lighting that is more realistic, this approach significantly slows down the ray-tracing based rendering process and requires environment map images. In order to generate large amounts of variable data we use random backgrounds from the SUN~\cite{xiao2010sun} scenes dataset instead.
\end{document}